\documentclass[acmsmall]{acmart}
\usepackage{amsmath,amsfonts}
\usepackage{algorithmic}
\usepackage{graphicx}
\usepackage{textcomp}
\usepackage{todonotes}
\usepackage{menukeys}
\usepackage{csquotes}
\AtBeginDocument{%
  \providecommand\BibTeX{{%
    \normalfont B\kern-0.5em{\scshape i\kern-0.25em b}\kern-0.8em\TeX}}}

\setcopyright{acmcopyright}
\copyrightyear{2018}
\acmYear{2018}
\acmDOI{XXXXXXX.XXXXXXX}

\acmJournal{JACM}
\acmVolume{37}
\acmNumber{4}
\acmArticle{111}
\acmMonth{8}

\begin{document}

\title{Human-Robot Team Performance Compared to Full Robot Autonomy in 16 Real-World Search and Rescue Missions: Adaptation of the DARPA Subterranean Challenge*\\
}

\author{Nicole Robinson}
\email{nicole.robinson@monash.edu}
\orcid{0000-0002-7144-3082}
\affiliation{%
  \institution{Monash University}
  \streetaddress{18 Alliance Lane}
  \city{Clayton}
  \state{Victoria}
  \country{Australia}
  \postcode{3800}
}

\author{Jason Williams}
\email{Jason.Williams@csiro.au}
\orcid{0000-0002-2416-075X}
\affiliation{%
  \institution{CSIRO}
  \streetaddress{1 Technology Court}
  \city{Pullenvale}
  \state{Queensland}
  \country{Australia}
  \postcode{4069}
}

\author{David Howard}
\email{david.howard@csiro.au}
\orcid{0000-0002-5012-7224}
\affiliation{%
  \institution{CSIRO}
  \streetaddress{1 Technology Court}
  \city{Pullenvale}
  \state{Queensland}
  \country{Australia}
  \postcode{4069}
}

\author{Brendan Tidd}
\email{brendan.tidd@csiro.au}
\orcid{0000-0002-7721-7799}
\affiliation{%
  \institution{CSIRO}
  \streetaddress{1 Technology Court}
  \city{Pullenvale}
  \state{Queensland}
  \country{Australia}
  \postcode{4069}
}

\author{Fletcher Talbot}
\email{fletcher.talbot@csiro.au}
\orcid{0000-0001-7845-4641}
\affiliation{%
  \institution{CSIRO}
  \streetaddress{1 Technology Court}
  \city{Pullenvale}
  \state{Queensland}
  \country{Australia}
  \postcode{4069}
}

\author{Brett Wood}
\email{brett.wood@csiro.au}
\orcid{0000-0002-2818-920X}
\affiliation{%
  \institution{CSIRO}
  \streetaddress{1 Technology Court}
  \city{Pullenvale}
  \state{Queensland}
  \country{Australia}
  \postcode{4069}
}

\author{Alex Pitt}
\email{alex.pitt@csiro.au}
\orcid{0000-0001-7501-6275}
\affiliation{%
  \institution{CSIRO}
  \streetaddress{1 Technology Court}
  \city{Pullenvale}
  \state{Queensland}
  \country{Australia}
  \postcode{4069}
}

\author{Navinda Kottege}
\email{navinda.kottege@csiro.au}
\orcid{0000-0002-2286-776X}
\affiliation{%
  \institution{CSIRO}
  \streetaddress{1 Technology Court}
  \city{Pullenvale}
  \state{Queensland}
  \country{Australia}
  \postcode{4069}
}

\author{Dana Kuli\'c}
\email{dana.kulic@monash.edu}
\orcid{0000-0002-4169-2141}
\affiliation{%
  \institution{Monash University}
  \streetaddress{18 Alliance Lane}
  \city{Clayton}
  \state{Victoria}
  \country{Australia}
  \postcode{3800}
}

\renewcommand{\shortauthors}{Robinson et al.}
\begin{abstract}
Human operators in human-robot teams are commonly perceived to be critical for mission success. To explore the direct and perceived impact of operator input on task success and team performance, 16 real-world missions (10\,hrs) were conducted based on the DARPA Subterranean Challenge. These missions were to deploy a heterogeneous team of robots for a search task to locate and identify artifacts such as climbing rope, drills and mannequins representing human survivors. Two conditions were evaluated: human operators that could control the robot team with state-of-the-art autonomy (Human-Robot Team) compared to autonomous missions without human operator input (Robot-Autonomy). Human-Robot Teams were often in directed autonomy mode (70\% of mission time), found more items, traversed more distance, covered more unique ground, and had a higher time between safety-related events. Human-Robot Teams were faster at finding the first artifact, but slower to respond to information from the robot team. In routine conditions, scores were comparable for artifacts, distance, and coverage. Reasons for intervention included creating waypoints to prioritise high-yield areas, and to navigate through error-prone spaces. After observing robot autonomy, operators reported increases in robot competency and trust, but that robot behaviour was not always transparent and understandable, even after high mission performance. 
\end{abstract}


\begin{teaserfigure}
    \includegraphics[width=\textwidth]{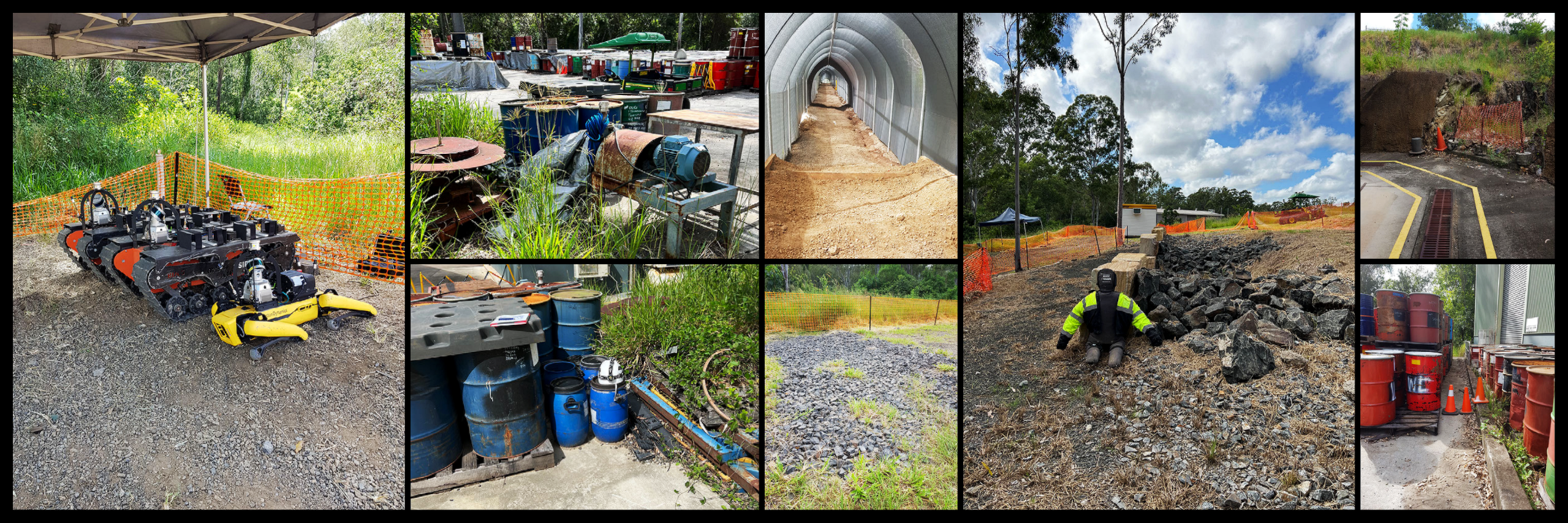}
    \caption{Experimental Setup for Testing Human-Robot Team Performance Compared to Full Robot Autonomy in 16 Real-World Search and Rescue Missions}
    \Description{Human-Robot Team Performance Compared to Full Robot Autonomy in 16 Real-World Search Missions}
 \end{teaserfigure}

\maketitle

\section{Introduction}
Autonomous robots combine numerous subsystems together such as sensing, navigation, localisation, planning, and control, to enable platforms to carry out tasks without human intervention. More capable autonomous robots can therefore become effective teammates that work alongside humans in a human-robot team scenario. For example, autonomous robots assisting their human teammates to achieve a mission-related goal~\cite{bauer2008human}. To achieve success in human-robot teams, effective teamwork between humans and robots is essential. Effective teamwork involves the balance between the need for close operator supervision and full independent robot autonomy without any oversight on their actions~\cite{bauer2008human,doi:10.1080/00140139.2018.1441449}. Therefore, human-robot teams often involve a task-load split between humans and robots that best suits the task. Heterogeneous robot teams often provide performance gains over homogeneous teams in relation to different capabilities to contribute to the task, which offers new opportunities for robots to contribute in beneficial ways under different levels of risk tolerance~\cite{9484733,Murphy2016}. In turn, operators provide their own unique strengths to the team, such as the capacity to conduct high-order goal planning and decision-making with incomplete information. In unison, human-robot teams can work together to overcome challenges that humans and robots alone are unable to do, helping to accelerate the utility and impact of human-robot teams to translate into real-world outcomes~\cite{michaelis2020collaborative}. In the human-robot team process, it is not always clear what the most suitable level of autonomy/supervision is to create successful human-robot teams, and what is the direct and specific benefit that human operators provide to human-robot teamwork above robot autonomy alone. To investigate such a research question involves the completion of tasks that both include and exclude human operators into the process.   

In this paper, we present a series of experiments to assess the impact of human operators working with state-of-the-art robot autonomy compared to robot autonomy alone in a set of outdoor field deployments. The experiment scenarios are modelled on the DARPA Subterranean (SubT) Challenge using the CSIRO Data61 Team as a test case~\cite{DBLP:journals/corr/abs-2104-09053}. The objective of the SubT challenge is to cover a large volume of unknown terrain to find as many artifacts as possible, which simulates a search mission using a robot team in the aftermath of a natural or industrial disaster to locate survivors~\cite{DBLP:journals/corr/abs-2104-09053}. We conducted a set of 16 real-world search missions over a total of 10\,hrs paired with detailed data analysis of team performance scores to assess the impact and involvement of the operator in mission-related outcomes. This experimental set evaluated human-robot team performance on key mission-relevant metrics such as total map coverage, artefacts found, and safety-related events, showing how human-robot teams compare to fully autonomous operation alone, and what advantages operators offer when they are able to intervene and direct the mission. 

\section{Background}

In this section, we will briefly review pertinent background literature in human-robot teaming, focusing on human-robot teaming in search and rescue. We will then review the format and scoring fo the DARPA Subterranean Challenge, which is used as the framework of our study.

\subsection{Human-Robot Teams with Robotic Teammates} 
Robots that can sense, navigate, localise and plan in an effective human-robot team configuration can contribute to beneficial mission-related outcomes in human-robot team scenarios~\cite{bauer2008human,6697830}. Human-robot teams have been tested across several different domain types for their effectiveness and capability to contribute to task success. Examples includes within urban search and rescue expeditions~\cite{doi:10.1177/1541931215591051, moonlight} and space exploration~\cite{hambuchen2021review}. 

Robot teammates that make intelligent and effective decisions on their own can help to extend the capacity and reach of the human involved in the team, otherwise referred to as the human operator. Robot teams are often under the direct control of a human operator, either through the direct teleoperation of robot movements, or by supervising autonomous robots to execute the task or action~\cite{liu2013robotic}. The operators role and level of involvement also can widely vary, depending on autonomy level in their robot teammates, which can span anywhere from full teleoperation all the way through to infrequent or brief involvement with the robot~\cite{vagia2016literature}. For example, robots within human-robot teams that continue on the initial actions set by the operator to then explore new regions, traverse more ground, or coordinate together in an autonomous way to assist the operator's mission directive~\cite{moonlight,lewis2010choosing,Murphy2016}. The human operator is often spending the most amount of time on direct teleoperation tasks, especially when the robot teammates are not able to make their own decisions. Instead, intelligent robot behaviours can help to reduce operator workload to allow the operator to focus on more urgent or pressing tasks during the mission, such as to focus on more important tasks, outcomes, or other team members~\cite{8169072,liu2013robotic}. More effective operator time can be critical to the mission, given that operators are often unable to directly control more than one robot at a time during complex tasks. Instead, operators can better control 4-8 robots that are acting in a semi-autonomous way~\cite{4651073}. Human-robot teams are showing notable promise for future applications, but the role of the operator for the level and type of involvement in the task can be important for team success~\cite{bauer2008human,doi:10.1177/0018720817743223,moonlight}. As robot teammates become more capable to contribute to mission-based outcomes, the role of the operator can transition into a more supervisory role rather than direct robot control~\cite{6697830}. Human operators can offer a strong sense of foresight, contextual awareness and higher-level prioritisation to ensure that the most critical and urgent tasks are addressed first during exploration~\cite{6697830,liu2013robotic,moonlight}. 

In real-world missions, operators can often spend their time assessing the robots' current state, and combining visual information provided by the robots to update their own view of the situation and environment to determine the next steps in the task~\cite{moonlight}. Where communications links permit, operators can assist by providing guidance or teleoperation to avoid critical incidents such as the robot becoming stuck, slipping, or colliding with objects, or re-directing the robot away from exploring areas that have already been well covered~\cite{1307409}. To operate the team, operators often process a large volume of information related to the mission, including robot status updates, team-related errors, multi-agent coordination, robot navigation choices and trajectories, human-robot team task allocation, communications links, environmental conditions and topography, key objectives for the search and rescue mission, as well as additional mission constraints such as total time~\cite{moonlight}. Therefore, operators must have access to operator control tools and interfaces that can allow the operator to build up sufficient awareness of the situation that the robot team is currently experiencing~\cite{6005237}. Human operators can also control more than one robot in the team, which can create even greater complexity with coordination between multiple robot viewpoints, functionalities, capabilities, and level of technical skill required for each task~\cite{4651073}. Due to the increased complexity of the supervisory task, human operators can also inadvertently contribute to negative outcomes during the mission. Operators are often affected by high cognitive load demands when working and supervising multiple robots, which can directly influence mission and task performance~\cite{6926363}, as well as increased cognitive load when operators are required to monitor more robots~\cite{doi:10.1177/1555343411409323}. For example, human operators contribute to more than 50\% of robot failures~\cite{Murphy2016,murphy2014disaster}, and operators that attempt to control too many robots in a single team can reach a clear limit on human-robot team operation~\cite{8913876}, eventually leading to deterioration in team performance~\cite{4651073}. While operators play a clear role in directing robots to achieve better outcomes, operators can also inadvertently contribute to performance errors and interruptions. Therefore, it critical to understand how to best utilize operators in the loop, and where operator intervention could best be used to minimise cognitive load while maximising mission outcomes. 

\subsection{Human-Robot Teams for Search and Rescue}
Disaster response and search and rescue is an area of human-robot teams in which humans and robots can work together to find as many survivors as possible without risking the lives of emergency personnel~\cite{Murphy2016}. Human-robot teams have been utilized to assist in human recovery after disaster-related events, such as at the World Trade Center bombing, La Conchita mudslide, and Hurricane Charley~\cite{doi:10.1177/154193120504900347}. Search and rescue missions that are led by human-robot teams often focus on covering as much ground as possible in an attempt to find the largest number of survivors, ensuring that emergency personnel can make informed decisions based on the most relevant and available information about the event~\cite{Murphy2016}. In the context of search and rescue, a mission outcome can involve directing the robot enter a hard-to-reach environment and create the next task set to explore additional areas to better understand the environmental layout to increase the success rate of finding survivors. Human-robot teams for search and rescue can help to protect and coordinate rescue personnel to reduce the need to enter dangerous and hazardous zones, minimising the risk of physical harm to people~\cite{Murphy2016}. Robots can also provide real-time data about the scenario to help emergency personnel to get critical information from hard-to-reach places, such as to take images of the location to send back to operators for their review and action~\cite{Murphy2016,moonlight}. Robot teammates to support tasks in disaster response has been linked to better field performance, and has helped to assist operators to complete their mission objective~\cite{yancodarpa2013}. To date, human-robot teams have often been tested used detailed simulations which often involve elements of search and rescue tasks~\cite{10.1145/1349822.1349825,6251723,hong2019investigating}. Human-robot teams in simulated tasks were reported to have located a higher number of victims, covered a larger area~\cite{6251723}, total scene exploration time and task performance compared to semi-autonomous and teleoperation modes~\cite{hong2019investigating}. There is a continued need to further explore the utility, improvement and deployment challenges related to human-robot teams in search and rescue tasks, including with real world testing outside of simulation-based tasks.

\subsection{DARPA Subterranean Challenge}
\label{sec:subt}
Robotics challenges have been proposed and created to help accelerate the testing and development of human-robot teams, such as the DARPA Subterranean (SubT) Challenge. Our experimental protocol aims to recreate conditions encountered during the recently-concluded SubT Challenge, and carried out by one of the top teams from that event, Team CSIRO Data61 (a collaboration between CSIRO Data61, Emesent and Georgia Tech). The overall goal of the SubT challenge was to identify and locate the most artifacts to within 5\,m in a set of unknown courses, each of which presented a variety of different obstacles and challenge elements to overcome. These challenges were designed to push teams to consider heterogeneous teams of robots with a strong emphasis on sensing, autonomy, information exchange, and hardware robustness. Competing teams were required to build a human-robot team solution that would involve operators supervising robots to navigate tunnels with vertical shafts, tunnels with varying levels and narrow passages, expansive cave networks with diverse structures and caverns, as well as urban areas with expansive and challenging layouts. Challenge artifacts included a set of objects that required different detection modalities, such as a cube (visual and Bluetooth signatures), helmet, rope, fire extinguisher, drill, vent, gas (CO$_2$ concentration), backpack, cell phone (visual, Bluetooth and WiFi signature), and survivor (mannequin with a visual and thermal signature). 

All teams had approximately 12 months to develop, integrate and test their solutions for the final stage. The Team CSIRO Data61 solution (detailed in ~\cite{DBLP:journals/corr/abs-2104-09053}) provided a range of supervision options to the operator, including teleoperation, waypoint navigation, directed autonomy and full autonomy. The most effective mode of operation during the SubT challenge was found to be directed autonomy, where the system operates autonomously utilising a multi-agent task allocation system. In this mode, the operator can influence the autonomous operation, either by directly assigning tasks, or by applying geometric prioritisation regions, either within a particular spatial region or for paths that cross through a region (where the latter is particularly effective for prioritising exploration of spaces with a priori unknown extents). The user interface concepts are described in more detail in~\cite{chen_multimodal_2022}. The majority of a robots' time is spent performing autonomous exploration tasks. Other autonomous tasks include synchronising data (i.e., navigating towards the base until all data is uploaded to and downloaded from the base), and returning on low battery. Robots exchange data with each other, such that one robot can simultaneously execute another robot's synchronisation task as well as its own. Mapping data are exchanged and solved independently on each robot, such that any robot can continue an unfinished exploration task (e.g., a branch of a junction that was not followed) of any other robot. Tasks that can be manually generated include ``go to'' and ``drop communications node''. As with autonomous tasks, robots collaboratively bid on these tasks to determine the robot best-positioned and equipped to execute the task. Droppable communications nodes extend the communications range deeper into the subterranean environment.

The challenge was broken up into two phases: the circuit phase and the final phase. In the circuit phase, participating teams competed in three preliminary events that were approximately six months apart: tunnel systems, urban underground and natural cave networks. The mission time was limited to 60\,mins in the Circuit Events and the Final Prize Run of the Final Event, and 30\,mins in the Preliminary Rounds of Final Event. The cave circuit event was cancelled due to COVID-19, and Team CSIRO Data61 staged their own event in natural caves in Chillagoe, Queensland; this data is utilised in Experiment 1 below. The Final Prize Round was held at the Louisville Mega Cavern in Kentucky in September 2021. 


In each of the runs, there were limits on the number of artifact reports that could be submitted to the scoring server to discourage spurious reporting, obliging the teams to perform a thorough review of the detections before submission. In the circuit events, a maximum of 40 artifact reports were allowed with 20 artifacts hidden in the courses. At the final prize run, only a maximum of 45 artifact reports were allowed with 40 artifacts hidden in the course, creating a significant incentive to avoid spurious detections. Each artifact report sent to the DARPA scoring server by the operator at the base station consisted of the artifact class, and the location relative to the reference frame provided by fiducial markers on the starting gate of the course. If the artifact class is correct and the reported location has a Euclidean distance of less than 5\,m from the ground truth location of the artifact, one point is scored. The operator will be immediately notified whether the report scored a point or not. 

Team CSIRO Data61's base station operator interface evolved over the three year competition with various views and screens for operating the robot team as well as for reviewing and verifying the artifact reports before sending them to the DARPA scoring server. To avoid spurious artifact detections, all robot detected artefacts were reviewed by the human operator prior to submission. Each robot had a perception system consisting of a machine-learning based object detection and a lidar based SLAM solution to detect, classify and localise potential artifacts. Detection reports bundle the classification and localisation data together to the central operator station for review by the human operator. The operator reviews the detections as a list on the GUI, and can review both the classification image as well as localisation information marked on an updated map. The operator is tasked with reviewing the detection classification and localisation data to ensure they corroborate each other. If so, the operator then sends the reviewed detection report for scoring. An external scoring server receives scoring reports from the operator and will pass or fail the report resulting in a potential score increase. 

The automated detectors had to accept high false alarm rates in order to achieve adequate detection performance, due to the generalisation error of the detector, operating in the a priori unseen environment. However, flooding the base station with candidate artifact reports from multiple robots would overwhelm the operator reducing their effectiveness in reviewing and sending verified reports to the scoring server. In the final event, Team CSIRO Data61 used automated spatial tracking of detected artifacts onboard the robots to significantly reduce the number of candidate artifact reports sent to the base station while maintaining a low detection threshold/high false alarm rate. After achieving the equal top score at the Final Event, Team CSIRO Data61 came second on the tie-breaker criterion of time of last detection. Further detailed information about the SubT Challenge rules can be found in~\cite{Darparules}. 

\begin{figure}[h]
  \includegraphics[width=0.8\textwidth]{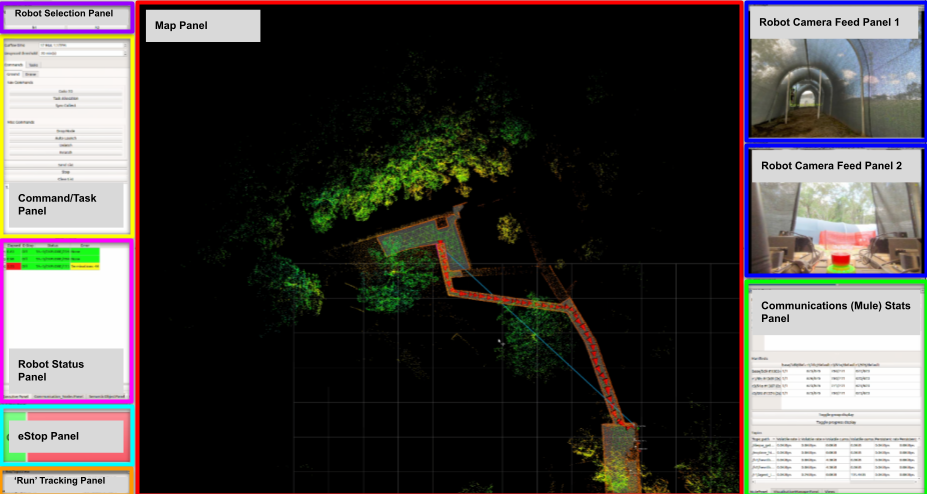}
  \includegraphics[width=0.8\textwidth]{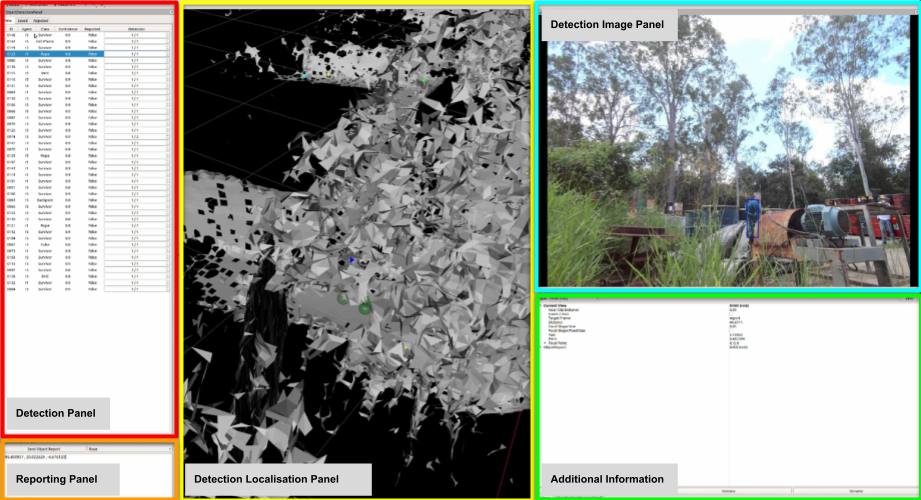}
  \caption{Team CSIRO Data61 Operator Interface. Top: GUI Commands. Bottom: Object Detection}
  \label{fig:interface}
\end{figure}

There is an excellent opportunity to further explore the role and contribution that operators provide to human-robot team performance, including when the task is related to a search and rescue mission. Such an experimental investigation would enable the close exploration of how different types of team performance metrics are achieved with and without the inclusion of a human operator in the mission, such as number of items found, unique distance travelled, and the total number of safety-related events.  

\section{Experiment 1: Operator Involvement and Performance in Four DARPA Subterranean Challenge  Runs (60\,min runs)}

To inform a controlled experiment, an initial analysis was retrospectively conducted on a pre-collected human-robot team run dataset to investigate the type, level of operator involvement and its impact on performance across four SubT Challenge runs that took place in a subterranean cave environment. No set hypotheses were proposed. The intention of this initial analysis was to explore the following research questions: 
\begin{enumerate}
    \item What type of operator intervention is often used? 
    \item How often does an operator intervene, and for what purpose? 
    \item What are the outcomes achieved by an intervention? 
    \item How does an intervention influence the mission score?
\end{enumerate}
The data analysed was Team CSIRO Data61's staging of a cave circuit event, in lieu of the formal DARPA event which was cancelled due to COVID-19. Details of the platforms and systems can be found in \cite{DBLP:journals/corr/abs-2104-09053}.

\subsubsection{Dataset Analysis and Results}
\begin{figure}[h]
  \includegraphics[width=0.8\textwidth]{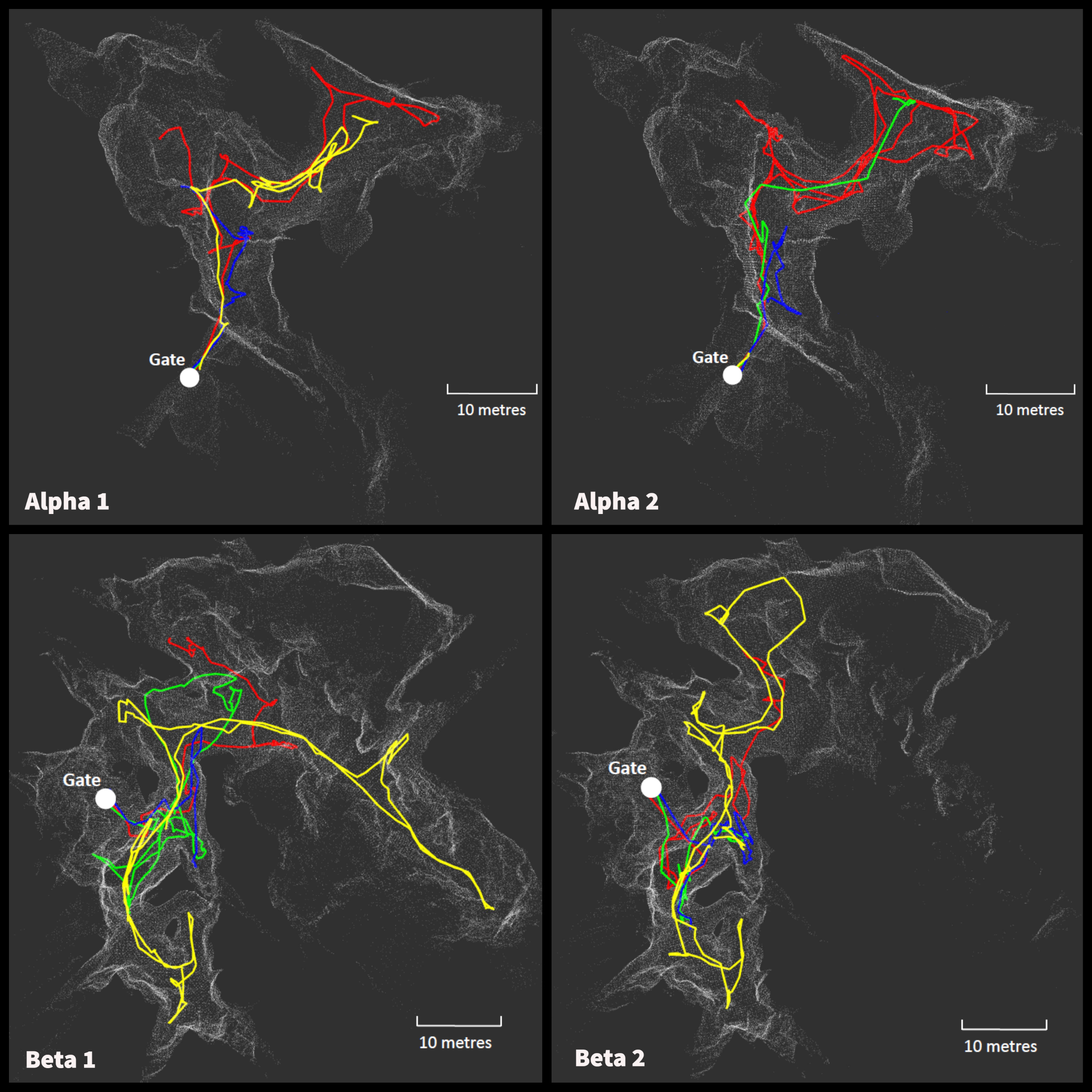}
  \caption{Course Map for four runs: Alpha 1, Alpha 2, Beta 1, Beta 2. The blue, red, green and yellow lines represent the trajectories of the robots. Yellow lines in Beta course represent drone trajectories; all other lines represent UGV trajectories.}
  \label{fig:prelimcoursemaps}
\end{figure}

The data consisted of team mission logs capturing robot state, mapping data, object detections and operator commands. This data was then replayed and processed offline to enable analysis. Analysis was conducted to identify operator involvement points through four course runs, which will be referred to as Alpha 1, Alpha 2, Beta 1 and Beta 2 (See Figure \ref{fig:prelimcoursemaps}). Initial results found there were four time-frames in which the operator intervened to control the robot team to explore new areas. A total of 16 out of 44 (36.36\%) artifacts were detected and reported correctly in the run: 14 (87.5\%) were detected by ground robots and 2 (12.5\%) were identified by the operator upon inspection of the map. The robot team travelled a total of 2,425\,m. Total intervention time via teleoperation was 8.65\,mins across all runs (3.6\%) with the intervention task to command the robots to explore other map areas. Ground robots were often more active during the first half of the run, which left the operator with sufficient time to go through the automated detection list and make the artifact reports, including to filter out most of the false alarm detection reports made by the robot team. When operators did intervene, their role was often to redirect the robots to new areas. Operator intervention in Alpha 1 and Beta 2 therefore resulted in beneficial mission-related outcomes for distance covered and artifact scoring. For instance, operator direction to explore new regions resulted in 2 additional artifact detections. All 4 runs had equivalent scores (4 points) with the Alpha 2 run having the lowest active time and travel distance to achieve the score. Operators had limited intervention (2\,mins on average), although involvement did contribute to improving mission-related goals, such as greater distance travelled and artifacts found. This analysis found that operators had very little involvement in directing the robots, suggesting that full robot autonomy for this task may be possible, and motivating a more detailed investigation into the role of the human operator. 

\section{Experiment 2: Testing Robot Autonomy Compared to Operator Involvement for Performance Metrics in Twelve Real-World Experimental Course Set Runs based on DARPA Subterranean Challenge (30\,min runs)}

Given the initial analysis from Experiment 1, a controlled experiment was designed to enable a clear comparison between human-robot team operation (i.e. operators working with a fully autonomous robot team) and full robot autonomy without human intervention. Experiment 2 was designed as a re-adaption of the SubT Challenge Final Event. In this experiment, we selected key team performance and mission-related metrics to identify the role and perceptions of the operator and evaluate human-robot team performance on human-machine team metrics~\cite{8404030}, such as total map coverage, number of found artifacts, and safety-related events, showing how human-robot teams compare to fully autonomous operation. Experiment 2 aimed to investigate performance-related impact as well as operator perceptions under two conditions:  operators could directly control a robot team with state-of-the-art robot autonomy (Human-Robot Team Condition, CH); this was compared to observing autonomous mission execution by the robots without operator input (Robot Autonomy Condition, CA). Note that, in the Robot Autonomy Condition, the operator was still responsible for reviewing and submitting the artefact reports. It is hypothesized that the Human-Robot Team mode (Condition H, CH) will outperform the Autonomous Exploration mode (Condition A, CA) on the following metric list:

\begin{enumerate}
    \item Higher final mission score for total number of found artifacts
    \item Greater distance and total course map coverage
    \item Fewer total number of safety-related events 
    \item Faster recovery time from error-related events
    \item Higher levels of cognitive load on the operators 
\end{enumerate}

\subsection{Human-Robot Team Composition}
\begin{figure}[h]
  \includegraphics[width=0.8\textwidth]{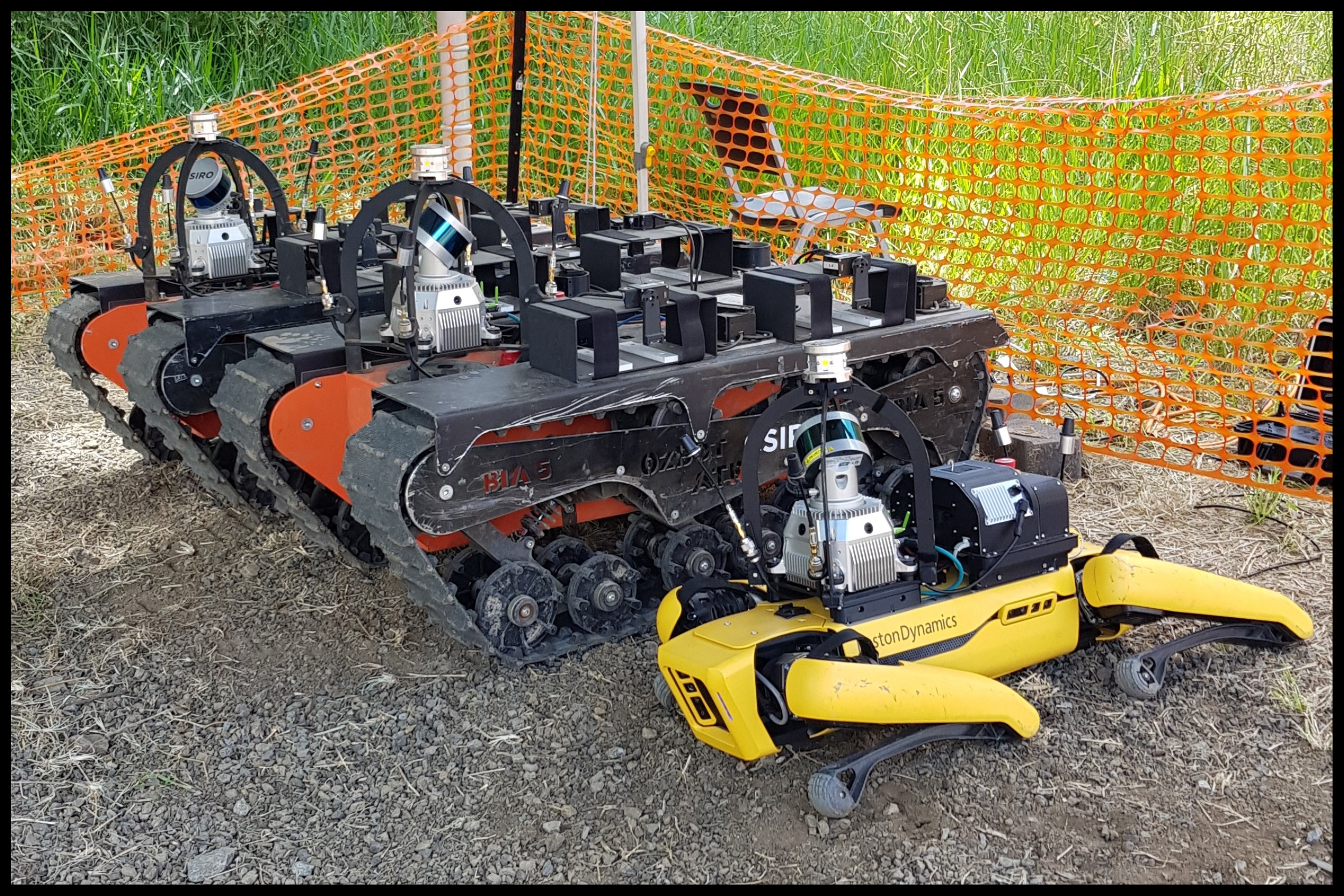}
  \caption{BIA5 All Terrain Robot (ATR) and Spot Robot in the Starting Gate Prior to a Course Run}
  \label{fig:atr}
\end{figure}

A single operator was asked to control a heterogeneous team of ground robots to find hidden artifacts in a set course outline, similar to SubT Challenge requirements. A total of four robots were available to use in the experiment: two BIA5 All Terrain Robots (ATRs) and two Spot Robots from Boston Dynamics. Nearly all runs were conducted with only three robots in each run (two ATRs and one Spot robot) with a single run using four robots for comparison purposes. The robot platforms can be seen in Figure \ref{fig:atr}. Due to complex considerations with communications node placement, these tasks are not generated automatically, and require operator initiation. For this reason, during this experiment, nodes were pre-positioned. Objects are scored if they are correctly identified and located to within an accuracy of 5\,m. This process is equivalent to that used at the SubT Challenge. In a small number of cases, the objects were not correctly positioned in the map in the scoring server; in these cases, failed scores were manually analysed and corrected in post-analysis. 

\subsection{Experimental Conditions} 
This experiment was conducted using a between-group research design for two conditions: Human-Robot Teams (CH) and Autonomous Exploration (CA):

\subsubsection{Autonomous Exploration Condition (CA)} This condition did not have any direct operator supervision of robot actions. To commence the run, the operator instructed the robots to a common starting point prior to being launched into autonomous exploration. After this event, the operator was not permitted to intervene; robots were followed by safety pilots, who would intervene only if the robot was about to encounter a high-risk condition or damage-related event. All robot autonomy choices were allowed to go ahead, such as if the robot was stuck, disorientated in its current location, or entering a segment of terrain  which the operator knew the robot would struggle to traverse. Operators were asked to confirm artifact detections provided by robots. As discussed in Section~\ref{sec:subt}, the generalisation error operating in an unknown environment necessitated operation with a high false alarm rate, and hence operator confirmation of autonomous detections was essential. In the Autonomous Exploration Condition, operators were asked to confirm artifact detections that were correctly identified and located. The operator was not permitted to correct errors in identity or location even if the images provided information that would allow that to be performed. This allowed for a fair comparison between conditions, to focus on operator control and robot autonomy. Time to human intervention was recorded for both conditions, including the time the robot first detected the artifact and an operator reviewed the detected image, as well as the time between when the operator reviewed the artifact report and scored it. However, it should be qualified by the fact that optimising this time was not part of the operational doctrine. Screen and audio recordings that were taken of the autonomous exploration runs were manually reviewed by an independent third party who did not contribute as an operator in the experiment to ensure that operators were scoring fairly across both conditions. 

\subsubsection{Human-Robot Team Condition (CH)} This condition allowed the operator to have full control over the robot team if they chose to intervene at any time, replicating the operator involvement allowed in the SubT Challenge. In addition to the functionalities described in the Autonomous Exploration Condition (CA), intervention actions included the ability to teleoperate the robot to specific locations, to modify the robots' waypoint or goal points, and to change the robots' intended exploration area, direction or task. 

\subsection{Course Preparation and Runs}
\begin{figure}[h]
  \includegraphics[width=0.8\textwidth]{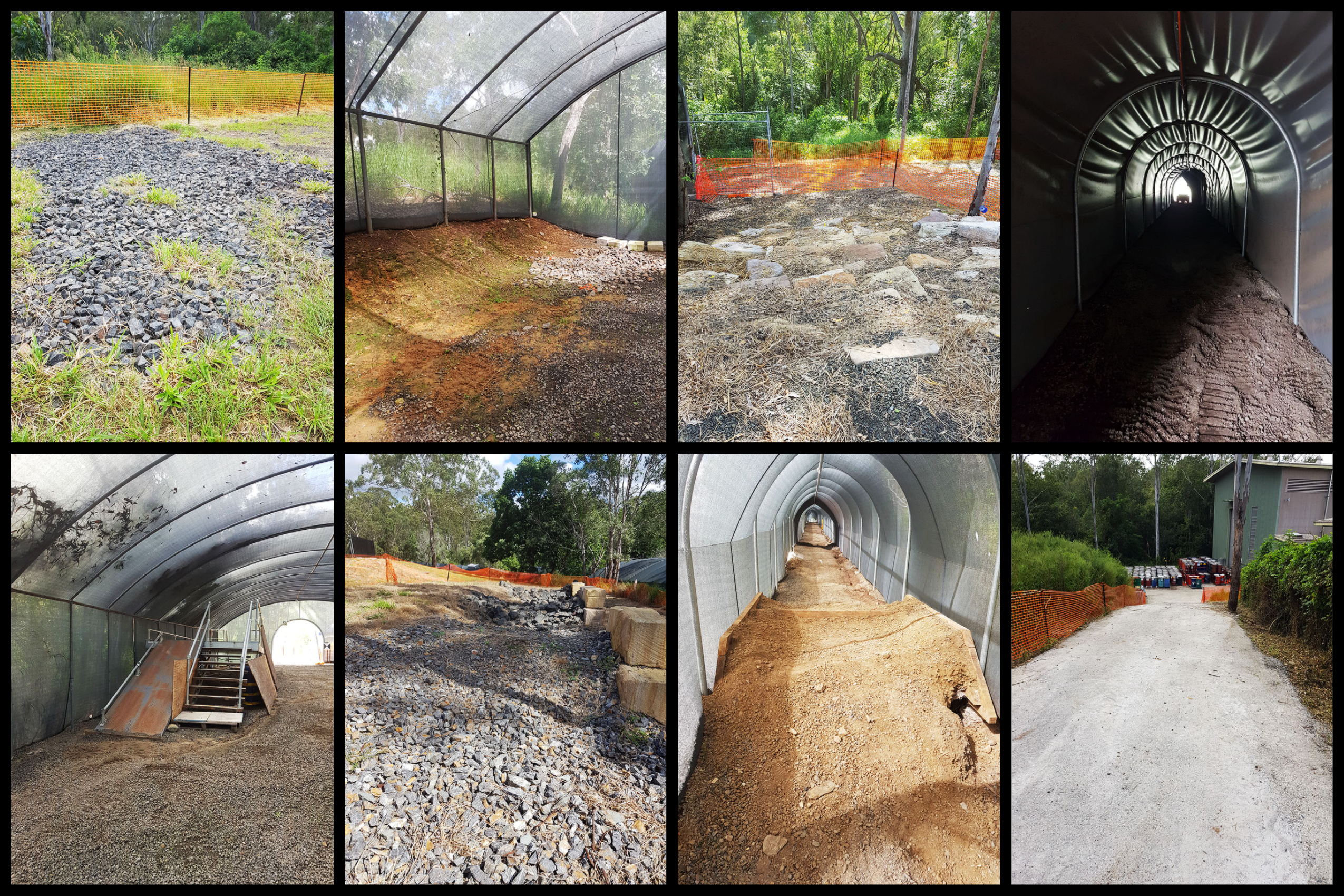}
  \caption{Terrain Challenges: Rocks, Ramps, Darkness, Loose Ground, Inclines, Declines, Stairs, Uneven Ground}
  \label{fig:terrainchallenges}
\end{figure}

\begin{figure}[h]
  \includegraphics[width=0.8\textwidth]{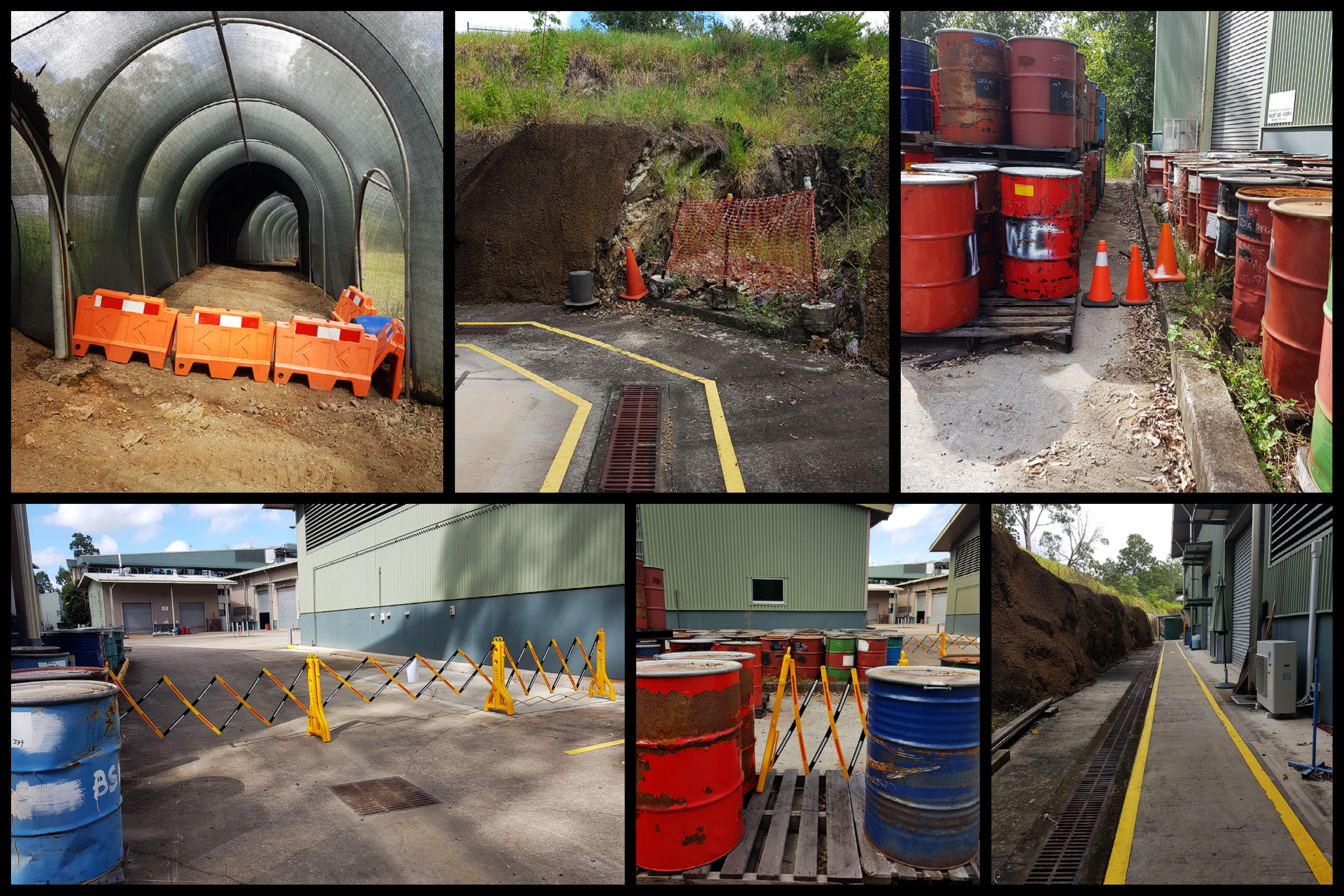}
  \caption{Human-Made Challenges: Roped Areas, Barrels, Barriers, Closed Doors, Building Components}
  \label{fig:humanchallenges}
\end{figure}

Each course run went for a total of 30\,mins at the CSIRO testing facility site in Brisbane, Australia. Twelve full course runs (also known as missions) were conducted over three sequential days. The testing schedule was conducted over three days to prevent hardware failures from other robot use influencing the experimental results, and to minimise software updates or changes influencing robot performance. Each operator was assigned a morning or afternoon session with a Human-Robot Team Condition (CH) run conducted first, followed by the Autonomous Exploration Condition (CA) run. The course was altered each day using temporary fencing, safety barriers and barrels to create more dynamic tunnels and pathways to explore, as well as dead ends that may or may not have an artifact (See Figure \ref{fig:terrainchallenges} and \ref{fig:humanchallenges}). Operators were not permitted to review or walk through the course before each trial. Each run contained a total of 16 artifacts with the artifact positions changing for each course variant. An automated system was utilised to keep track of the run score (i.e., the number of objects correctly detected within 5\,m of their ground truth location). First, a map is automatically generated by navigating the course with the robots, and aligning that map to the reference frame established by the global origin at the ``starting gate''. Subsequently, on each day as artifacts are placed, the artifacts are located in a prior map based on photographs (e.g., Figure \ref{fig:day1}) and entered into the automated scoring system. Since deployment of communication nodes was not automated, nodes were pre-positioned within the course to enable reasonable communications within the course bounds.

\subsubsection{Course Layouts}
\begin{figure}[h]
  \includegraphics[width=0.8\textwidth]{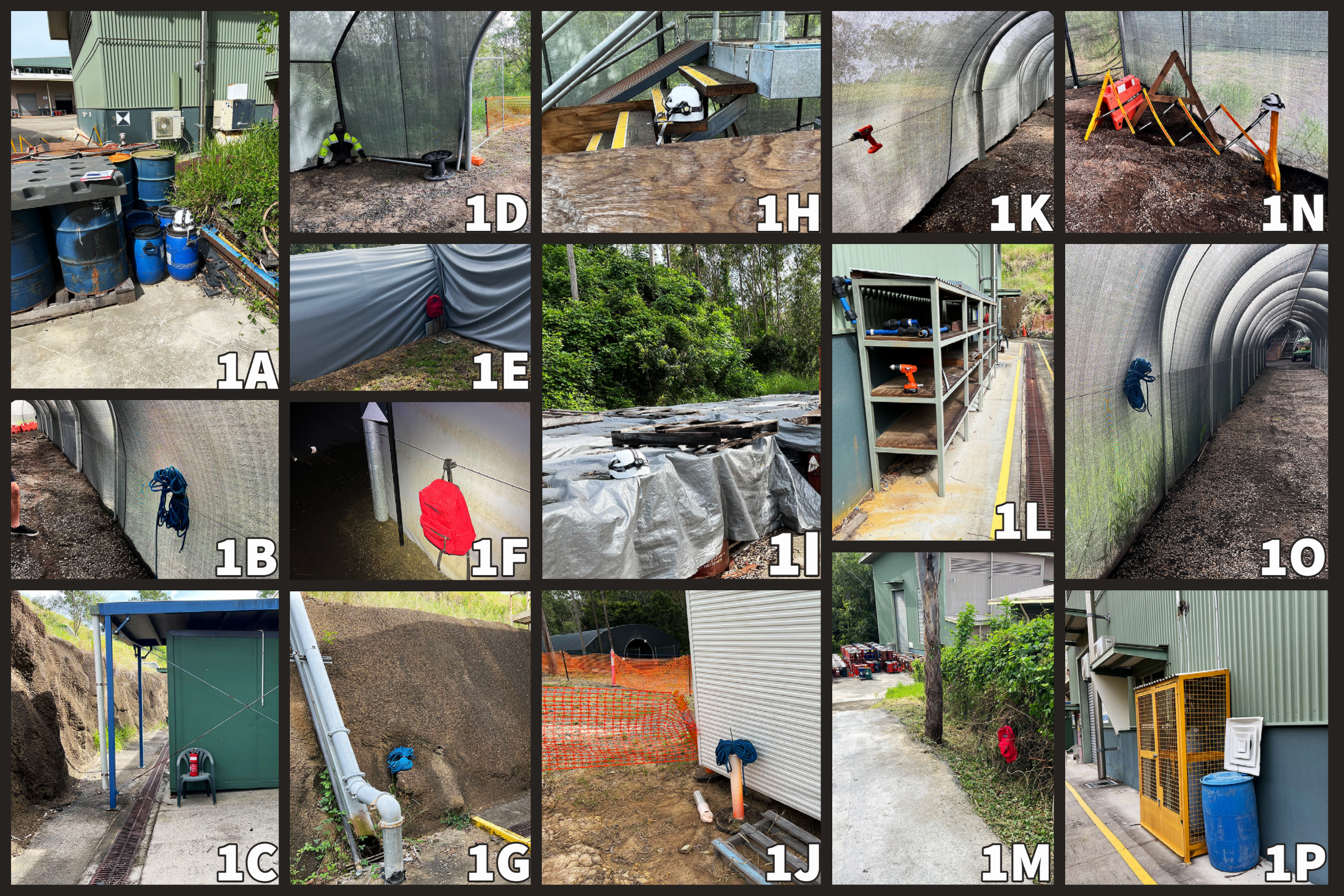}
  \includegraphics[width=0.8\textwidth]{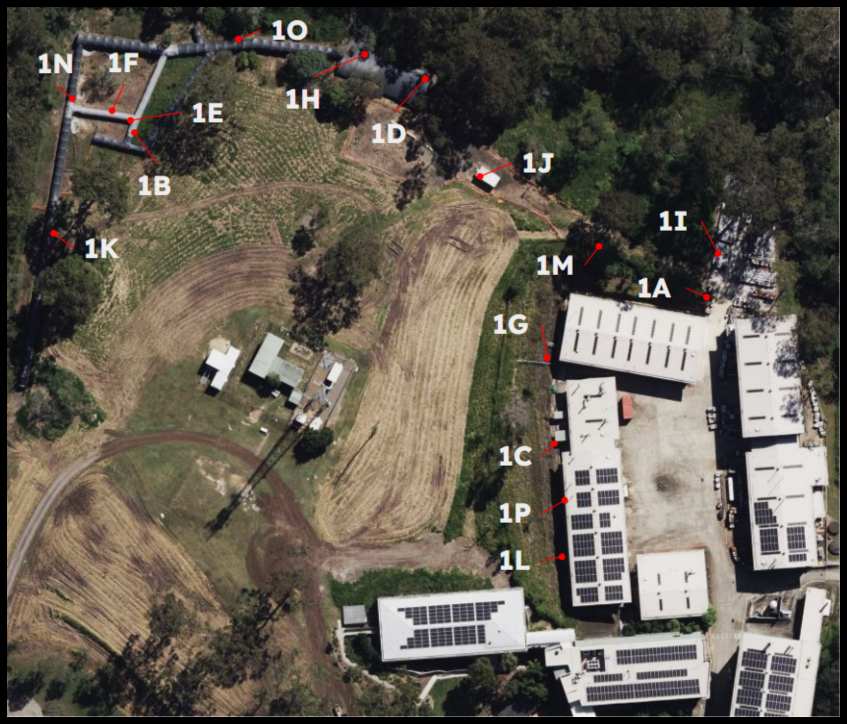}
  \caption{Course 1 Setup and Artifact Locations}
  \label{fig:day1}
\end{figure}

\begin{figure}[h]
  \includegraphics[width=0.8\textwidth]{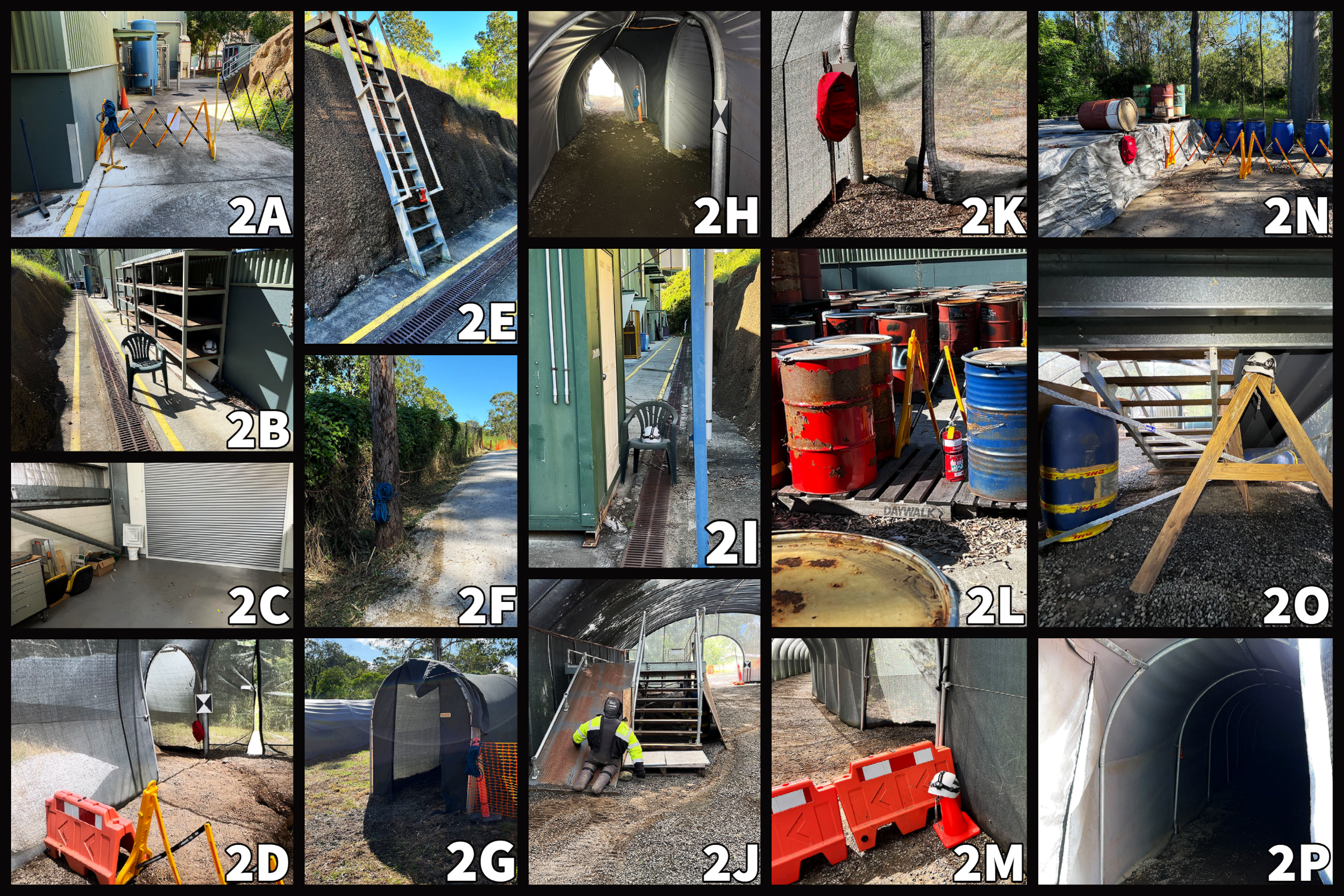}
  \includegraphics[width=0.8\textwidth]{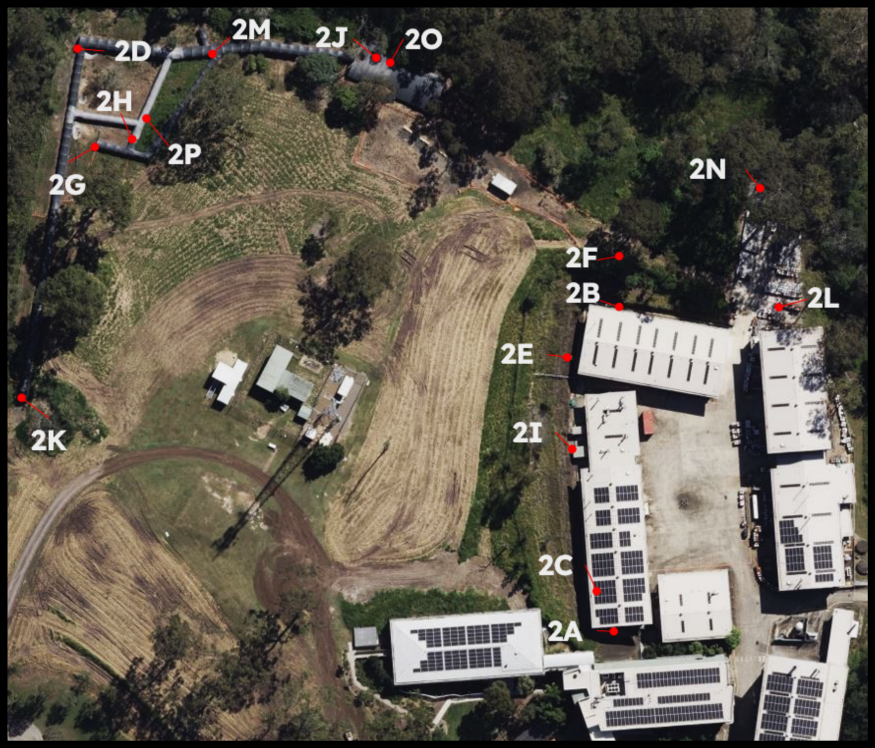}
  \caption{Course 2 Setup and Artifact Locations}
  \label{fig:day2}
\end{figure}

\begin{figure}[h]
  \includegraphics[width=0.8\textwidth]{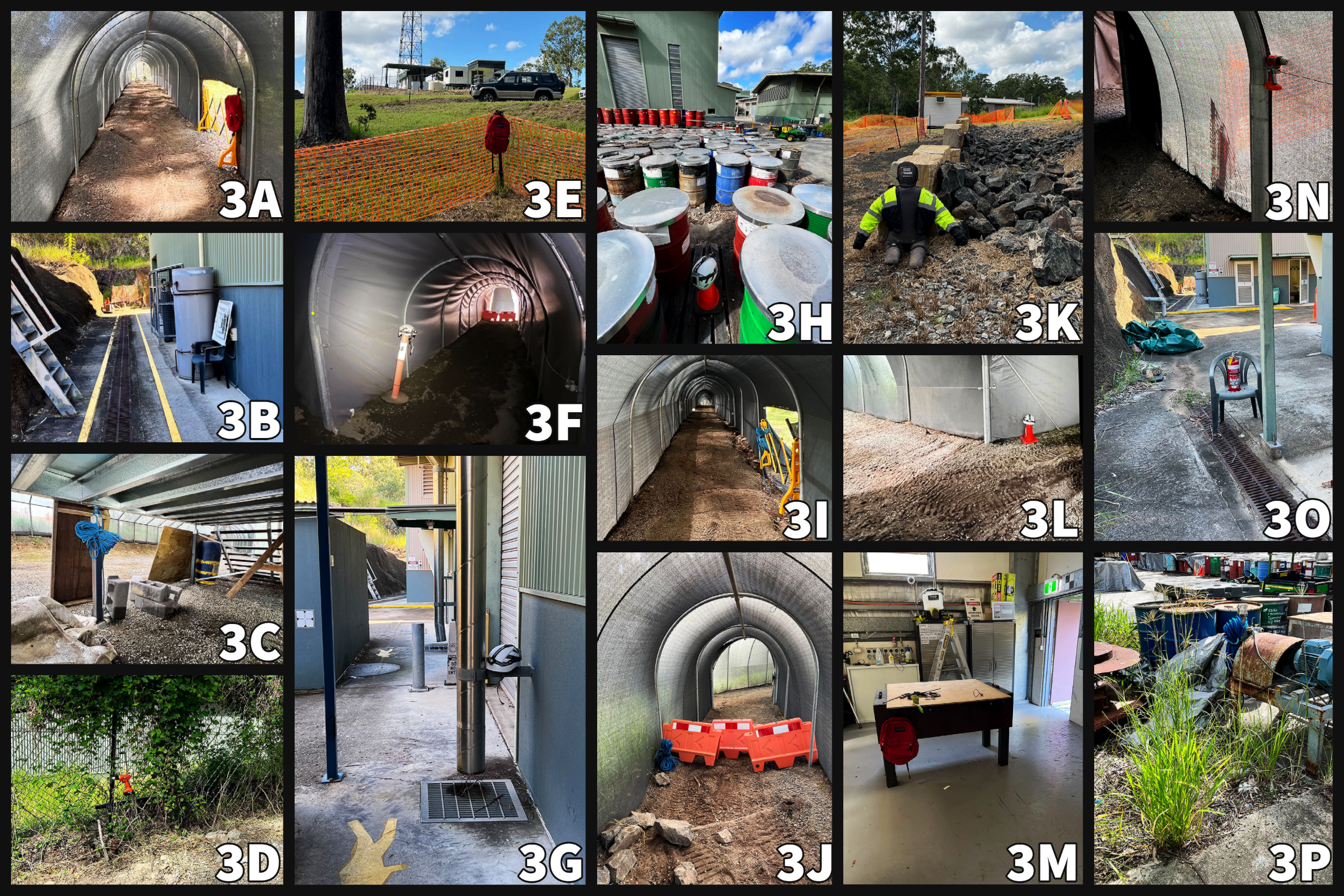}
  \includegraphics[width=0.8\textwidth]{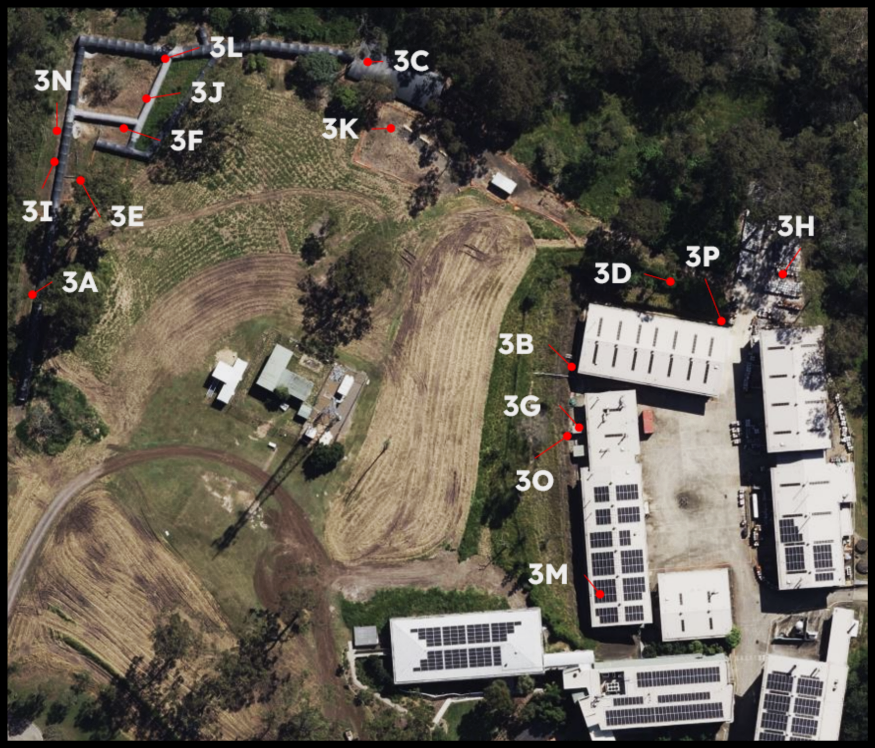}
  \caption{Course 3 Setup and Artifact Locations}
  \label{fig:day3}
\end{figure}

Course 1 consisted of four missions (mission 1-4). Course 1 in Figure \ref{fig:day1} was used for Missions 1-4. Mission 1 and 2 had three robots (2 x ATR and 1 x Spot) whereas Mission 3 and 4 used four robots (2 x ATR and 2 x Spot). There was a total of 16 artifacts: 4 helmets (A, H, I, N), 4 ropes (B, G, J, O), 3 backpacks (E, F, M), 2 drills (K, M), 1 fire extinguisher (C), 1 survivor (D) and 1 vent (P). Course 2 in Figure \ref{fig:day2} was used for Missions 5-8. There was a total of 16 artifacts: 4 ropes (A, F, G, H), 4 helmets (B, I, M, O), 3 backpacks (D, K, N), 2 drills (E, P), 1 vent (C), 1 fire extinguisher (L) and 1 survivor (J). Course 3 in Figure \ref{fig:day3} was used for Missions 9-12. There was a total of 16 artifacts: 4 ropes (C, I, J, P), 3 backpacks (A, E, M), 2 drills (D, N), 4 helmets (F, G, H, L), 1 survivor (K), 1 vent (B), and 1 fire extinguisher (O). 

\subsection{Procedure} 
All robots were thoroughly checked by the site team to ensure that each robot had sufficient power with no malfunction or errors before each run. All personnel not involved in the experiment were also asked to vacate the course. The operator was based in a demountable building in the middle of the course for the duration of the experiment. Inside the command centre, the operator was left to conduct the run with an experimenter present as a quiet observer for any assistance or information requests during the testing session. Each run was timed to have a duration of 30\,mins with clear start and stop time markers. Operators were given time markers of how long was left, including for 20, 15, 10 and 5\,mins. Time markers were given to operators if they requested additional information on the remaining time available to complete the run. All team members (operators, experimenters and safety observers) had radio communication set up for sharing critical course and run information, including potential hazards or challenges. Operators were permitted to communicate with the experimenter to ask for experiment-related information, such as remaining time. Operator utterances were also captured via audio recording, given their importance to describing the operators style, decision-making and planning~\cite{yancodarpa2013}, but operator utterance data was not analysed for this paper. Operators from different runs were not permitted to discuss course runs with each other and were encouraged to have minimal contact between runs, such as to remain off-site when they were not involved in the run to help minimise cross-contamination effects.

\begin{figure}[h]
  \includegraphics[width=0.8\textwidth]{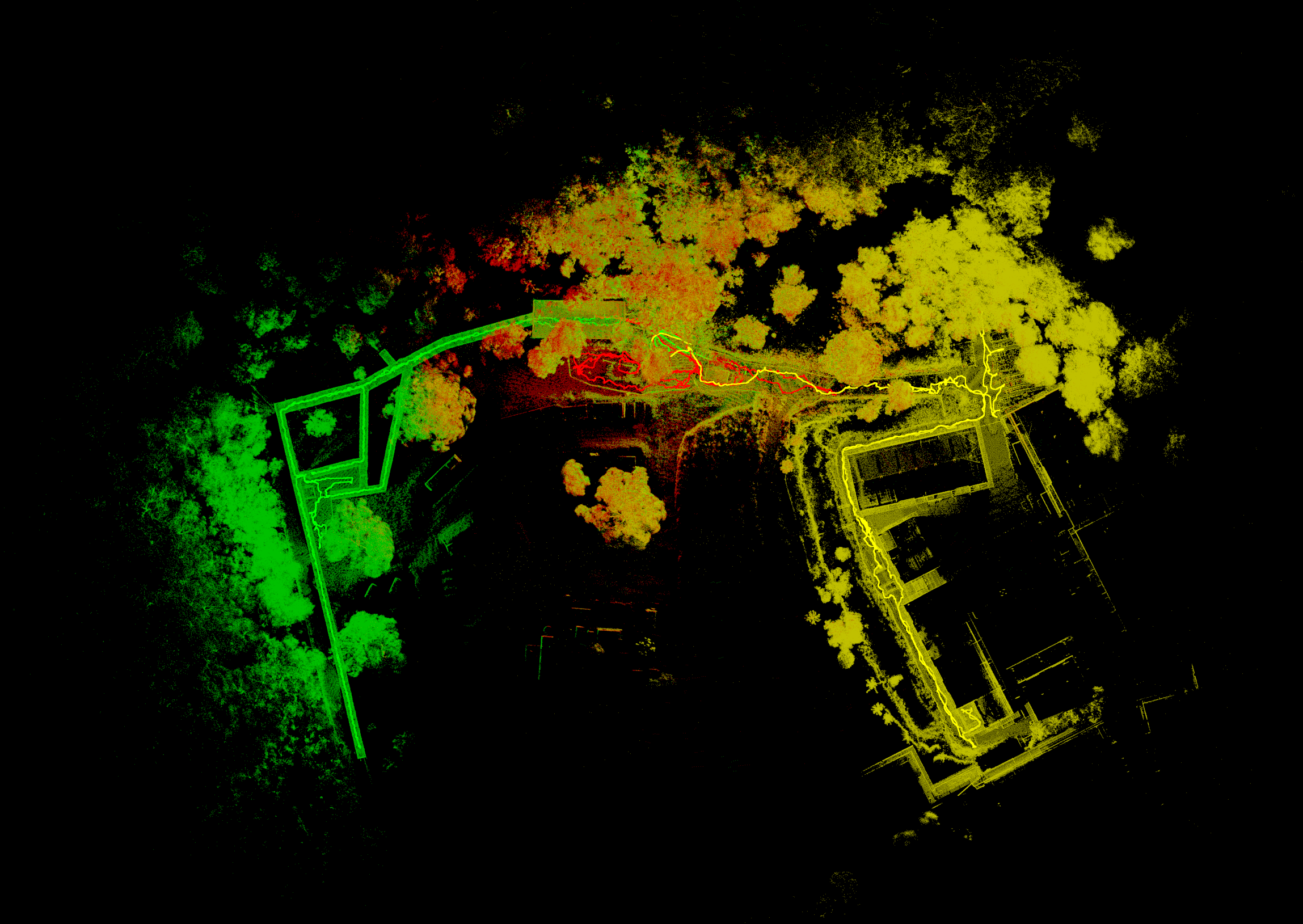}
  \caption{Example SLAM Map from an Autonomous Exploration run - Mission 12}
  \label{fig:slam}
\end{figure}

Once the experiment was ready to go, all robots were taken to the starting zone to be activated in the open-space area. In each run, the operator commenced by locating the gate in order to re-establish the reference coordinate system, in which object locations are reported.  Safety observers were allocated to each robot to follow at a suitable distance to observe, but not interfere, with the robot's current task. Observers were permitted to intervene via emergency stop (eStop) if the robot was going to damage itself, a building, object or person. Operator screens were recorded, including over-the-shoulder recordings for operator movements, communication and screen interaction. Operators wore a micro recorder with a lavalier microphone to comment on or narrate their current operation methods. Operators were instructed to maintain their natural communication method as they normally would during testing and were allowed to narrate of their actions and methods during their run if they wished to do so. Once the run was complete, communication was disseminated to all safety operators to monitor the robots as they returned to base. Once operators had finished all of their duties to relocate the robots back to the start position, operators were then prepared to start the qualitative interview segment. After the run, operators were asked to complete an assessment of task load for operator-assisted runs and autonomous exploration runs. Operators were then given the chance to request any further information or ask questions. See Figure \ref{fig:slam} for an example of the mission run course outline. 

\subsection{Qualitative Interview}
Operators completed a 10\,min semi-structured interview after each run. Interview questions were related to the course run and chosen to better understand operator performance, the level of involvement from the operator, and to collect more information on unexpected or error-related events during the course run. Examples included, but were not limited to, the following questions:

\begin{enumerate}
    \item Can you tell me how this course run went?
    \item How would you describe this course set up?
    \item How would you describe your performance as an operator? 
    \item What would you have done differently as an operator?
    \item Did anything unexpected or challenging occur during the run?
    \item Is there anything else you would like to comment on for this run? 
\end{enumerate}

Qualitative interviews were transcribed as strict verbatim pacific transcription via third party professional transcription services with ISO 9001 (Quality Management Systems) certification. All qualitative responses were stored, collated and analysed with NVivo Version 22. Manifest content analysis was conducted using a standardised qualitative method~\cite{elo2008qualitative, hsieh2005three, vaismoradi2013content}. In initial data preparation, interview responses were reviewed to classify emergent patterns and themes using initial notes and prospective data codes. In the organisation phase, an inductive approach was taken to create open code categories under relevant headings to prepare for subsequent analysis. In the reporting phase, clustered codes were assigned into a final set of categories and checked for accuracy and category allocation~\cite{elo2008qualitative, hsieh2005three, vaismoradi2013content}. All codes were checked, confirmed or reallocated to a more suitable category by the lead author to confirm the final category set. For clarity across the qualitative data analysis, select examples of robot terminology has been amended for continuity between the two operators, such as robot names and technical task names.

\section{Results - Quantitative Data}
In this section, composite and individual mission data will be presented to examine the differences between Autonomous Exploration (CA) and Human-Robot Team (CH) on human-robot team performance. 

\subsection{Operator Characteristics} 
Two expert operators were invited to participate. Selected operators were very similar in their level of training and familiarity with the SubT Challenge. Operator (O1 and O2) were males between 35-40 years of age with tertiary-level education in mechatronics. Operator 1 and 2 had both been the lead operators for approximately 50 runs and the non-lead operator for another 50 runs to make a total of 100 human-robot team runs and 100\,hrs of experience. Both operators had a total of 24-30\,months of experience with the system with both reporting a score of 9/10 for DARPA challenge knowledge, 9/10 for ATR operation and 9/10 for Spot operation. 

\begin{figure}[h]
  \includegraphics[width=0.45\columnwidth]{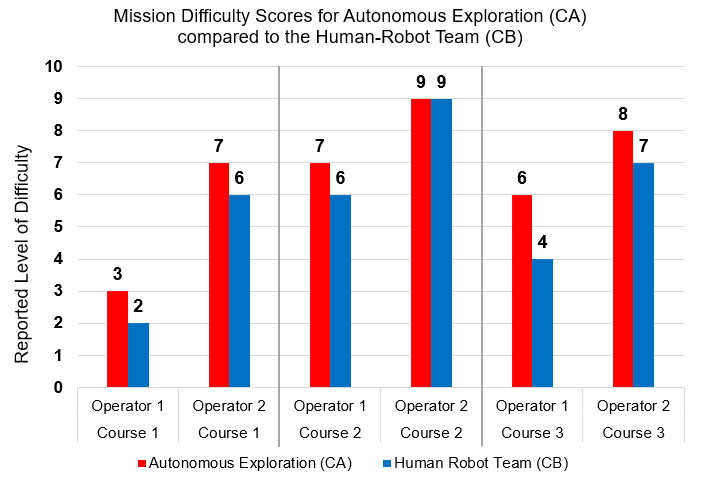}
    \includegraphics[width=0.45\columnwidth]{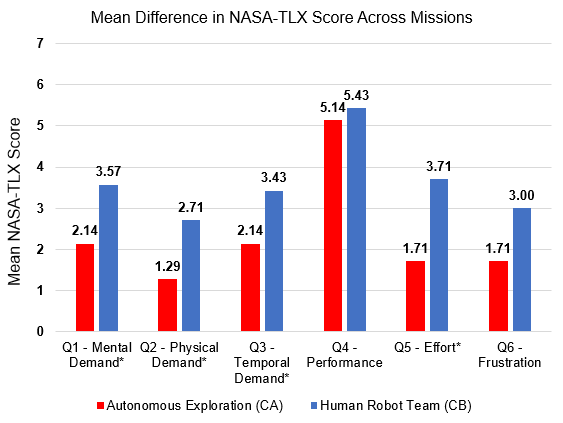}
    \caption{Operator Ranking of Course Difficulty for the Autonomous Exploration (CA) and Human-Robot Team (CH) Condition | Course Run Summary - NASA-TLX Scores. *$p =<.05$}
  \label{fig:operatorrating}
\end{figure}

\subsection{Operator Task Ratings, Task Load and Intervention Level}
On average, operators were involved in teleoperating the robot team for a total of 2\,mins and 3\,s (7.6\% of total mission time) in the Human-Robot Team missions (Range, 0:00 to 03:05). Except for Course 2 for Operator O2, both operators rated their perceived difficulty to complete the course higher for the Autonomous Exploration (CA) condition compared to the Human-Robot Team (CH) condition (See Figure \ref{fig:operatorrating}). Across 12 missions, both operators rated the mental demand, physical demand, temporal demand, and task effort to be significantly higher for operating the run during the Human-Robot Team (CH) condition compared to the Autonomous Exploration (CA) condition (See Figure \ref{fig:operatorrating}). There was no significant score difference in their perceived performance outcome between the Human-Robot Team (CH) condition compared to the Autonomous Exploration (CA) condition.

\begin{figure}[h]
  \includegraphics[width=0.8\textwidth]{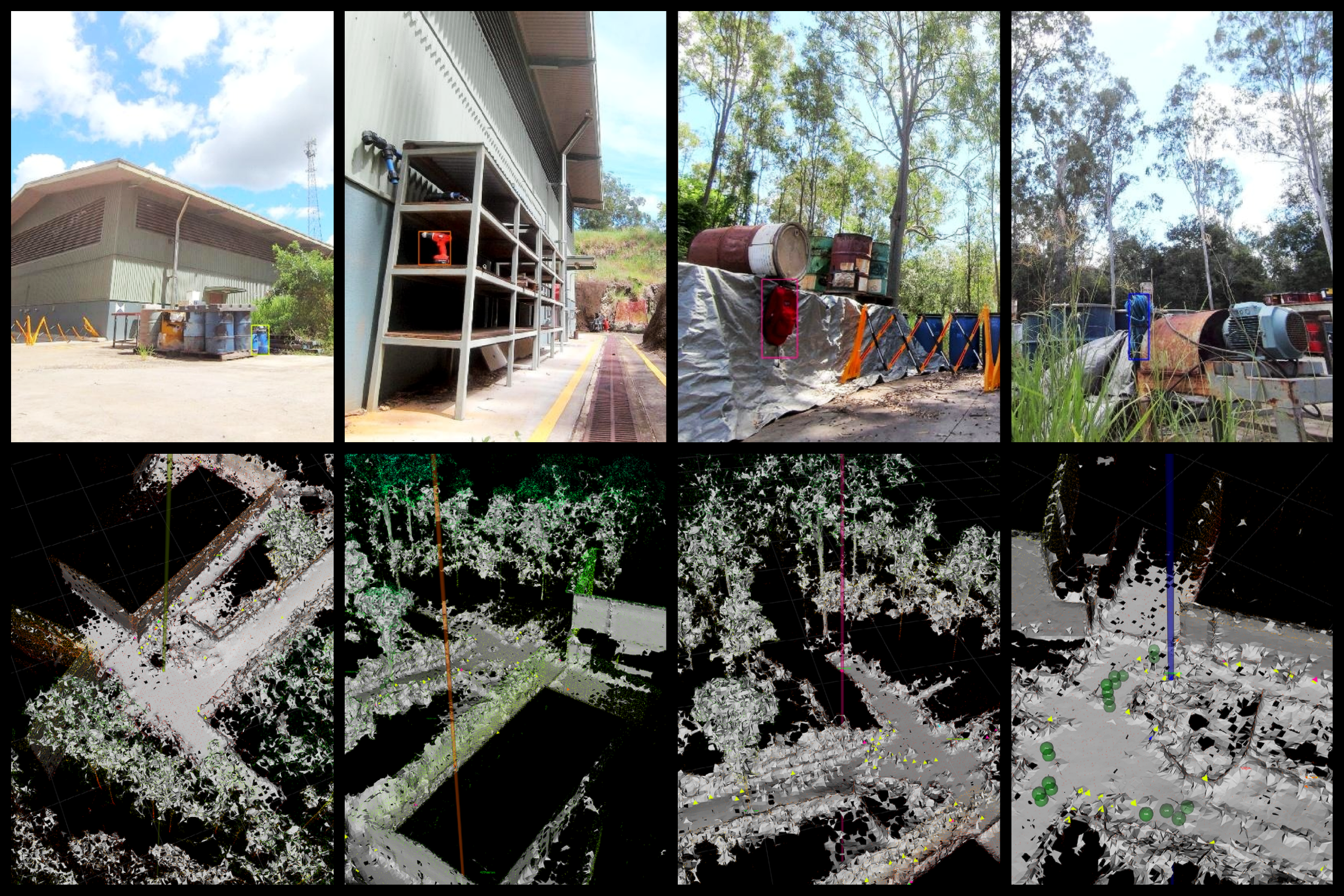}
  \label{fig:detect}
  \caption{Examples: Artifact Detection and Location}
\end{figure}

\subsection{Final Mission Score (Artifacts Found) and Course Run Times}
\subsubsection{Final Mission Score}
The Human-Robot Team (CH) condition scored 90 artifacts (94\%), and the Robot Autonomy (CA) condition scored 81 artifacts (84\%) out of a total of 96 artifacts (+10.52\%). In Mission 8, two artifacts (2H and 2I) were classified outside of the 30\,min run, and were therefore not scored. In reviewing matched mission pairs, there were four instances in which both conditions missed the same artifacts, three instances in which the robot missed an artifact that the human-robot team found, and one instance in which the human-robot team missed an artifact that the robot team found. Examples of detection images and their location can be seen in Fig~\ref{fig:detect}.

\subsubsection{Course Modes and Times} For course modes, there were four main categories reported for the experiment: eStop, teleoperation (teleop), directed autonomy mode led by the operators actions (directed) and autonomy mode without any operator directions or input (autonomy). Figure ~\ref{mode16} and ~\ref{mode712} demonstrates the mode types for each run with odd numbers representing Human-Robot Team runs, and even numbers representing Robot Autonomy runs. In the Human-Robot Team Condition, operators often used directed autonomy tasks during the mission run (70\% of mission time) compared to fully autonomous tasks (12\% of mission time) and teleoperation time (1.25\%).  

  \begin{figure}
  \includegraphics[width=\linewidth]{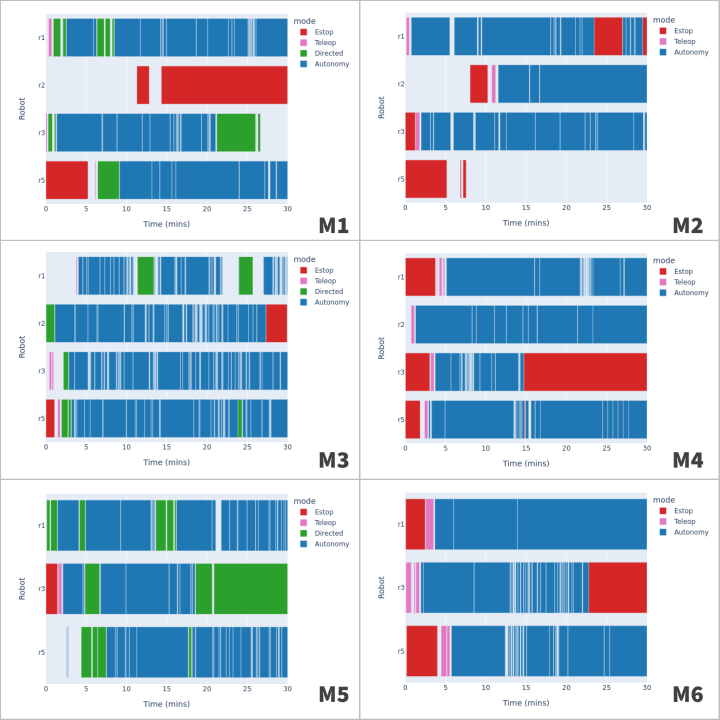}
  \caption{Robot Autonomy Mode for Mission 1 to 6}
  \label{fig:combinedtraj}
  \end{figure}

  \begin{figure}
  \includegraphics[width=\linewidth]{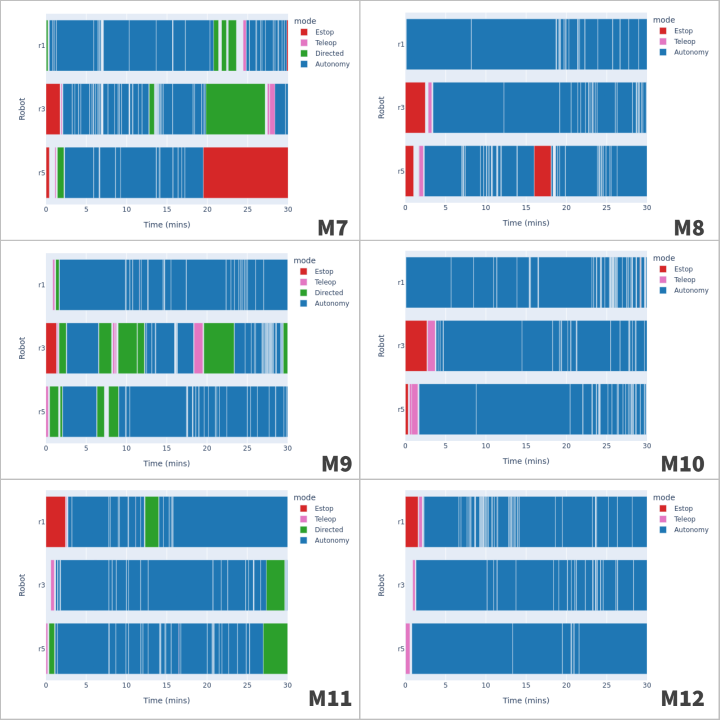}
  \caption{Robot Autonomy Mode for Mission 7 to 12}
  \label{fig:combinedtraj}
  \end{figure}

The Human-Robot Team condition was faster on average to obtain the first detection (2:24\,min) compared to the robot autonomy team (3:02\,min). The Human-Robot Team was slower in the average time between the robot first detecting the artifact and an operator reviewing the detected image (1:43\,min, CH) compared to the Robot Autonomy condition (1:08\,min, CA). The Human-Robot Team was slower in the average time between reviewing the artifact report and scoring it (1:41\,min, CH) compared to Robot Autonomy condition (0:31\,min, CA) when using list wise deletion. In Mission 1 and 2, the Human-Robot Team was 11:30\,mins faster to achieve 15 artifacts compared to the Robot Autonomy team with the same artifact total. However, in Mission 11 and 12 the Robot Autonomy team was 4:01\,mins faster to achieve 16 artifacts compared to the Human-Robot Team condition. 

\begin{figure}[h]
  \includegraphics[width=\linewidth]{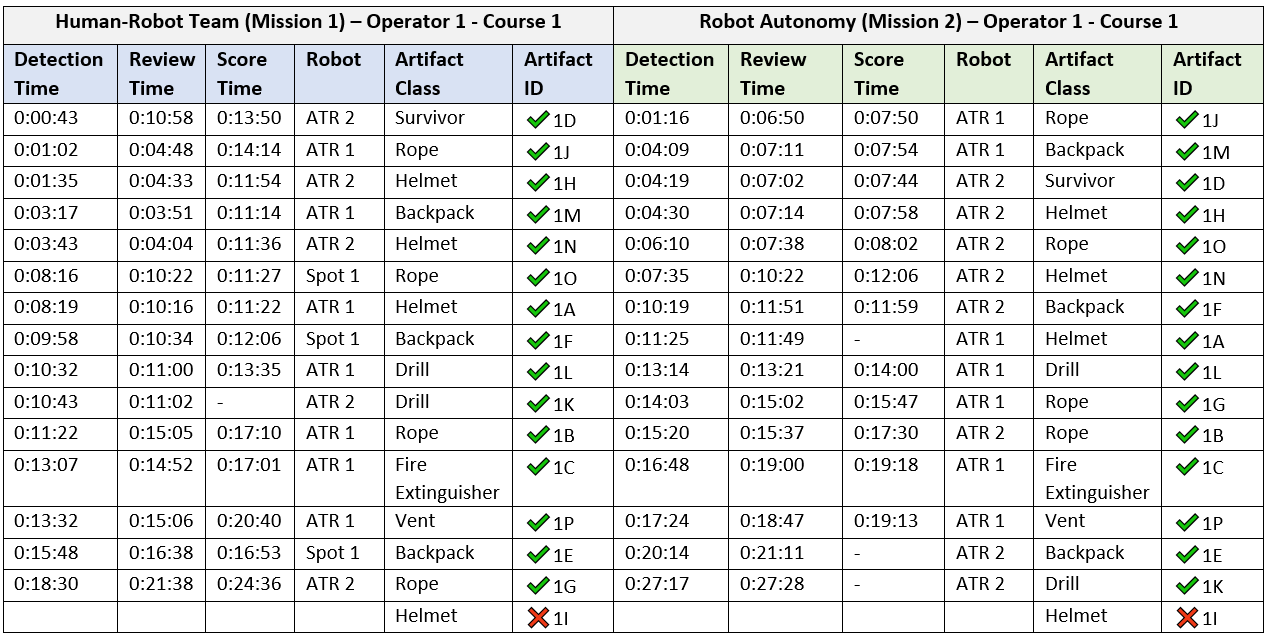}
  \label{fig:mission12}
  \caption{DARPA Scoring Results for Mission 1 to 2}
\end{figure}

\begin{figure}[h]
  \includegraphics[width=\linewidth]{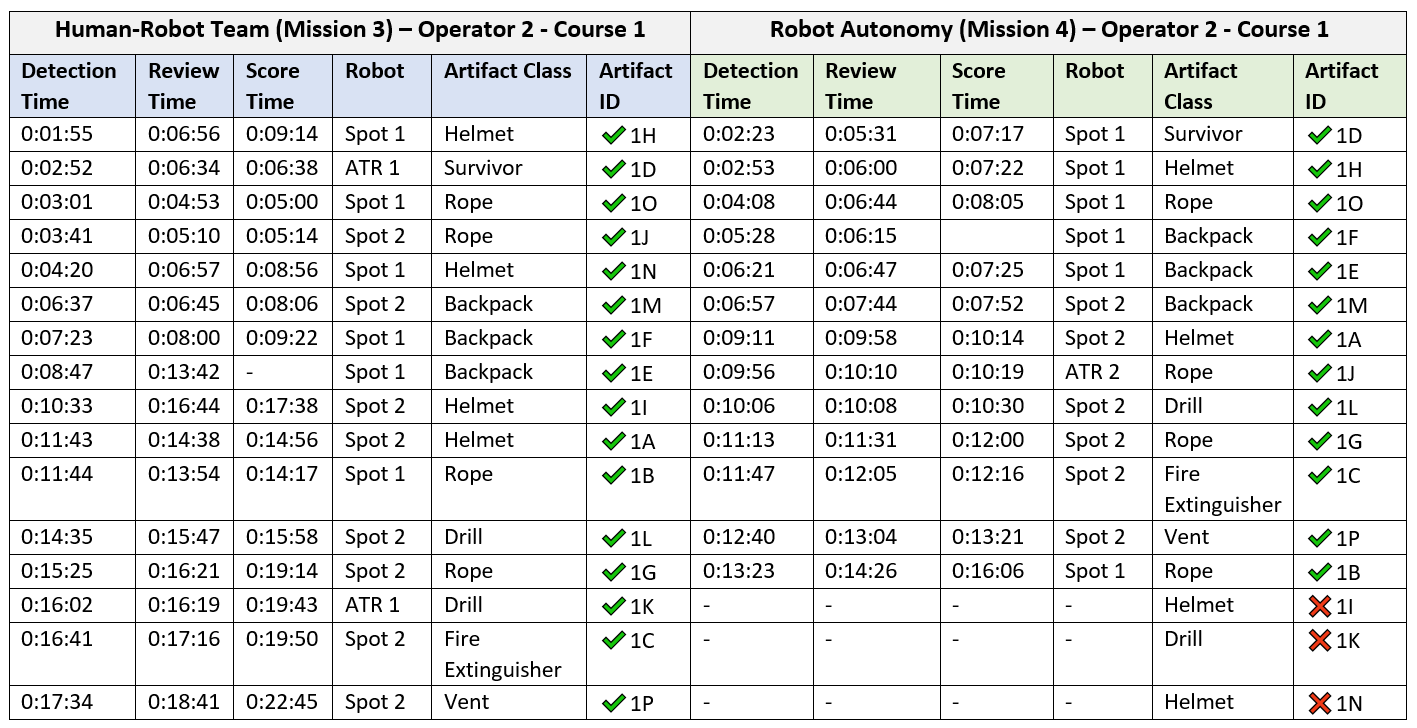}
  \label{fig:mission34}
  \caption{DARPA Scoring Results for Mission 3 to 4}
\end{figure}

\begin{figure}[h]
  \includegraphics[width=\linewidth]{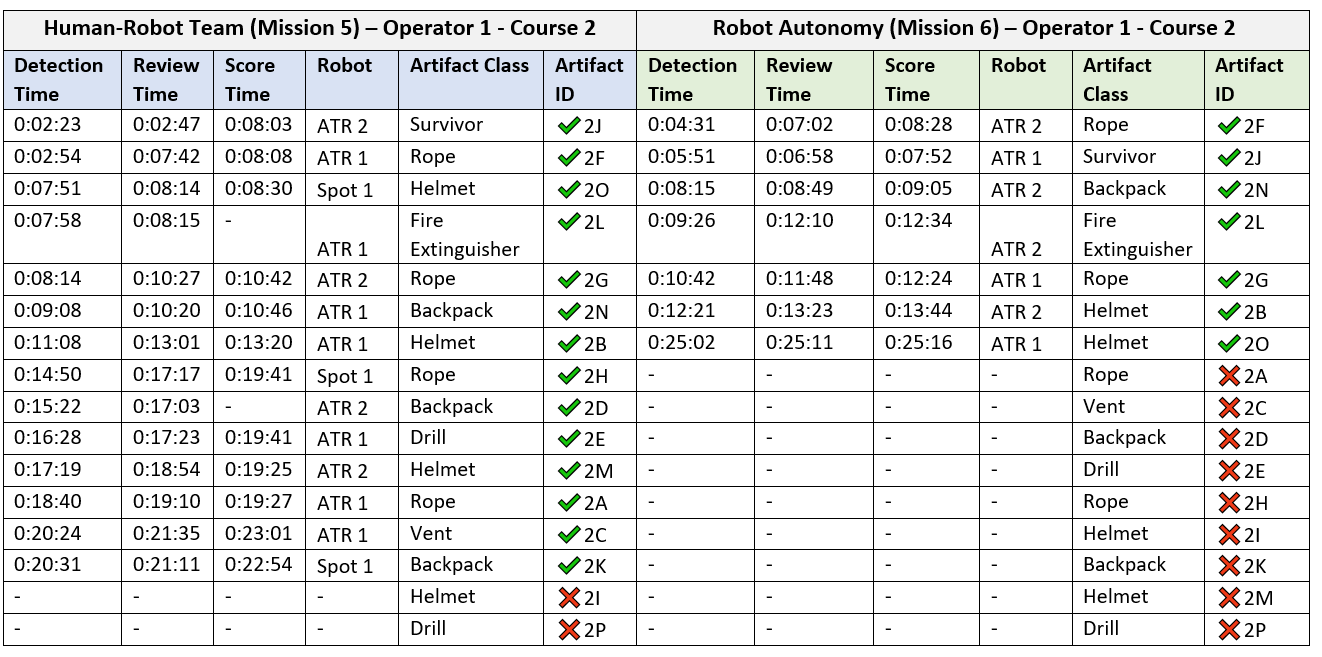}
  \label{fig:mission56}
  \caption{DARPA Scoring Results for Mission 5 to 6}
\end{figure}

\begin{figure}[h]
  \includegraphics[width=\linewidth]{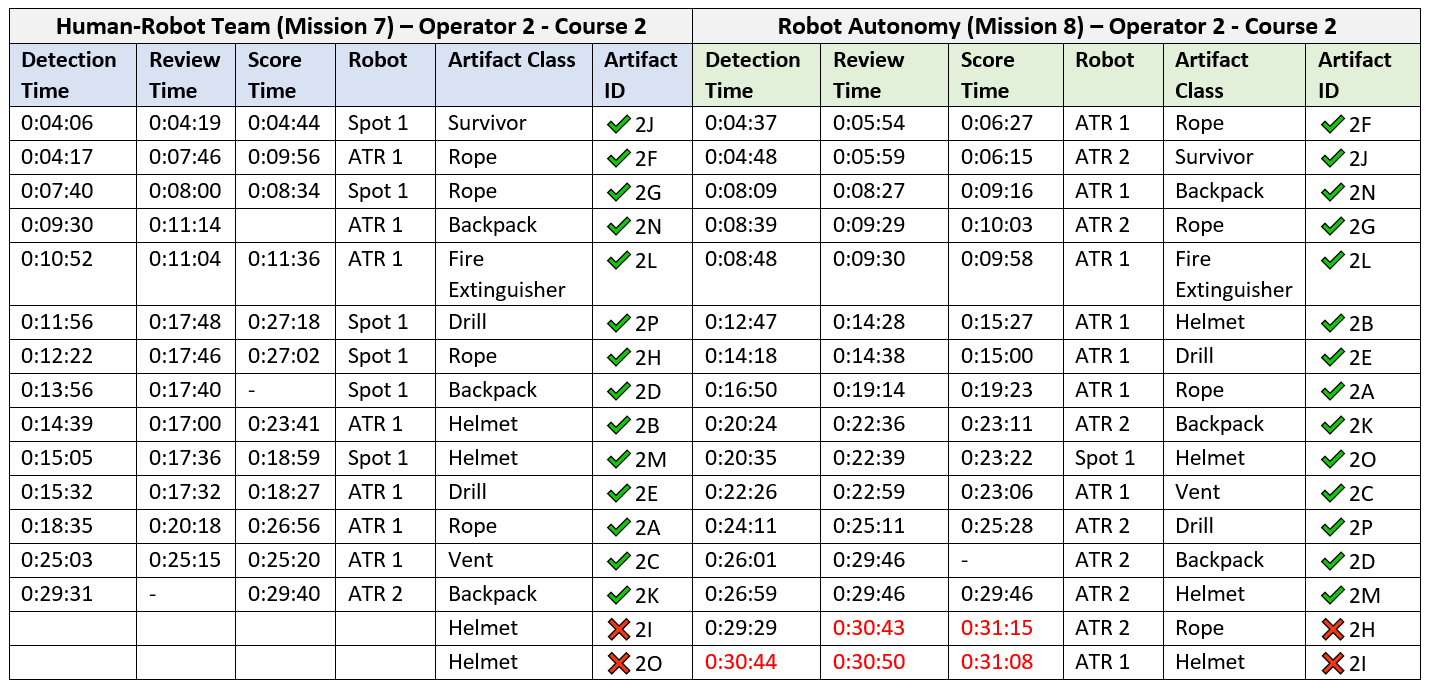}
  \label{fig:mission78}
  \caption{DARPA Scoring Results for Mission 7 to 8}
\end{figure}

\begin{figure}[h]
  \includegraphics[width=\linewidth]{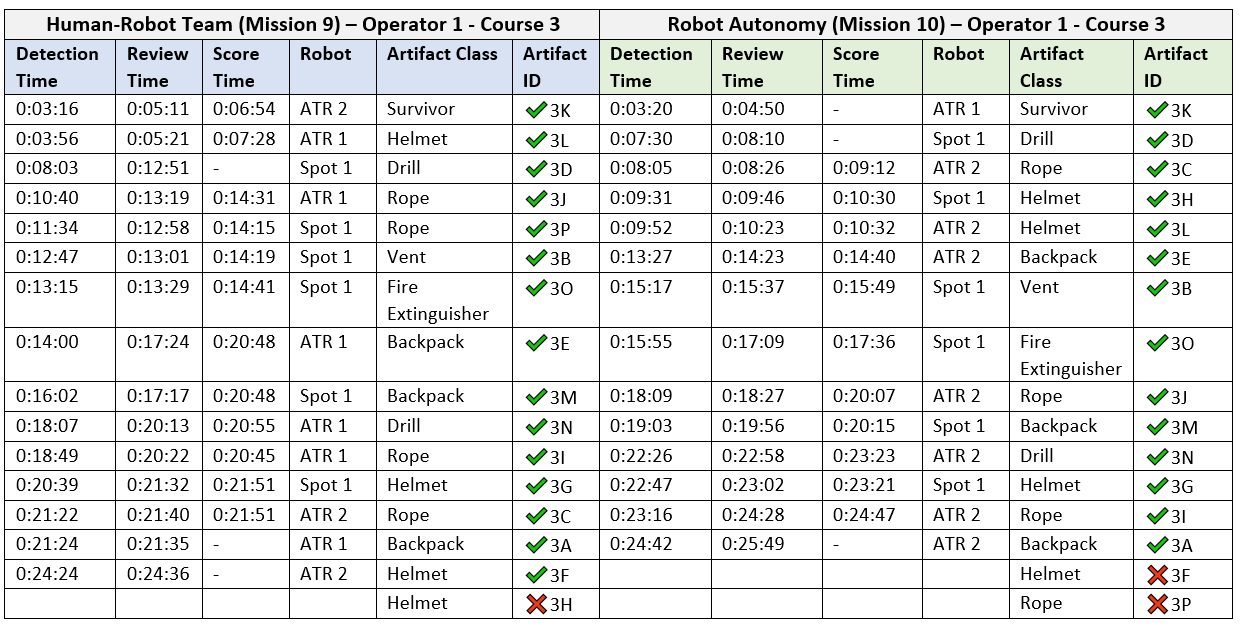}
  \label{fig:mission910}
  \caption{DARPA Scoring Results for Mission 9 to 10}
\end{figure}

\begin{figure}[h]
  \includegraphics[width=\linewidth]{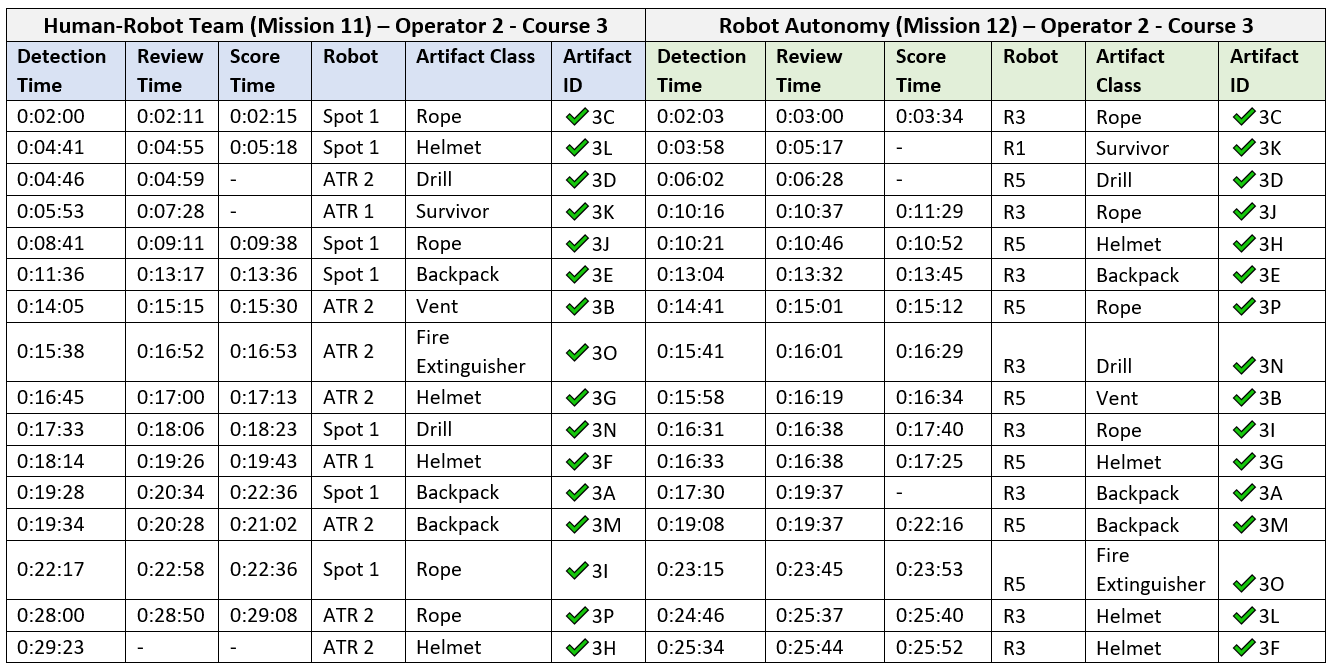}
  \label{fig:mission1112}
  \caption{DARPA Scoring Results for Mission 11 to 12}
\end{figure}

\begin{figure}[h]
  \includegraphics[width=0.9\columnwidth]{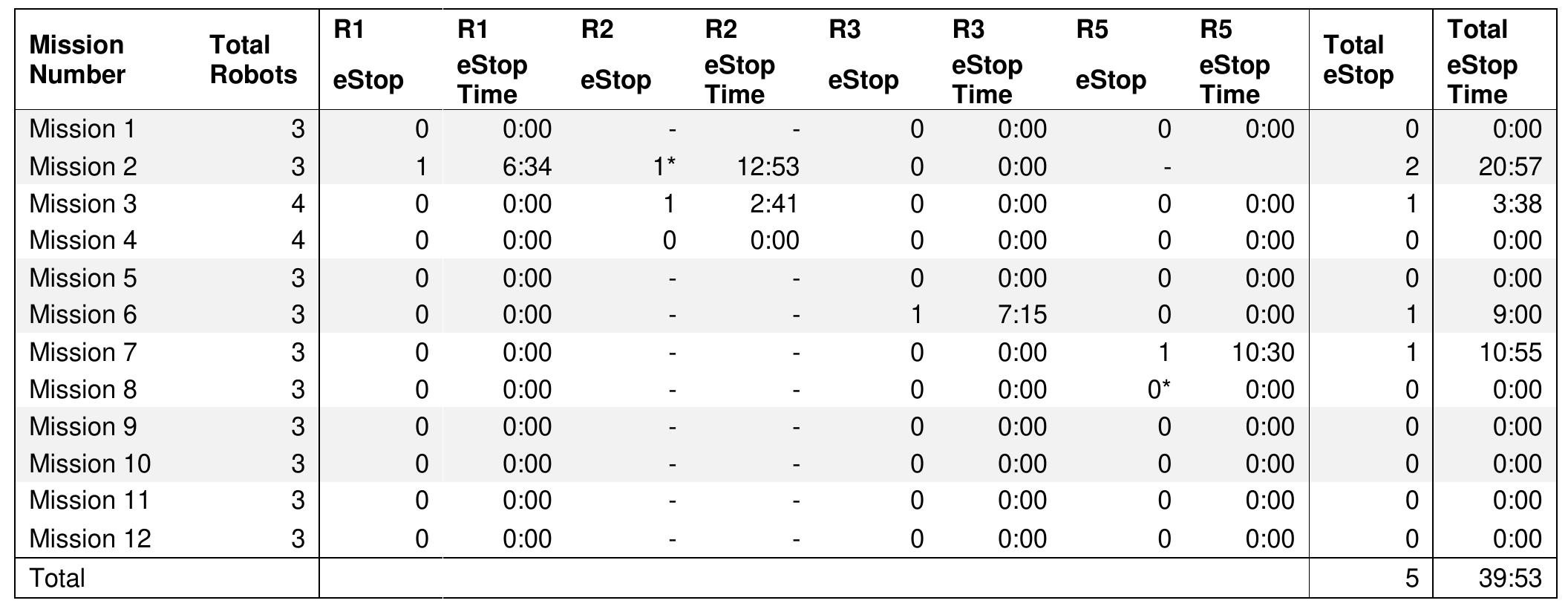}
    \caption{Totals for eStop Function for Safety or Error-Related Event. Asterisks mark exceptions as recorded in the text.}
    \label{fig:estop}
\end{figure}

\subsection{Total Course Map Coverage, Total Distance, Safety and Error-Related Events}
\subsubsection{Total Course Map Coverage and Total Distance}
Total distance and course map coverage was calculated using a composite total of meters squared covered by all robots for the full 30\,min runs. Total distance travelled was the full distance based on a composite score of each robot's distance travelled during each mission run, and total course map coverage calculated as the unique course coverage found by the full robot team. Human-robot team condition covered more unique ground (+1777.00\,m$^{2}$ or +5830.052\,sqft, +10.56\%) compared to robot autonomy (total coverage: 17713.75\,m$^{2}$, CH; 15936.75\,m$^{2}$, CA). The human-robot team achieved a higher total distance covered using total trajectory of all of the robots (+1130.38\,m or +3708.596\,ft, +12.71\%) compared to robot autonomy condition (total distance: 9454.78\,m, CH; 8324.40\,m, CA).

\begin{figure}[h]
  \includegraphics[width=0.45\columnwidth]{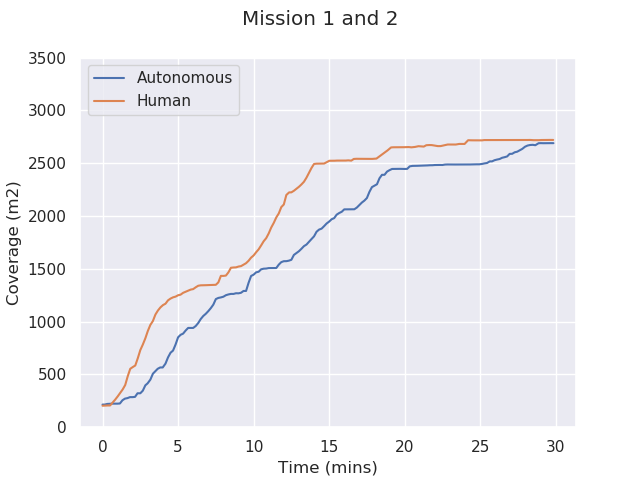}
      \includegraphics[width=0.45\columnwidth]{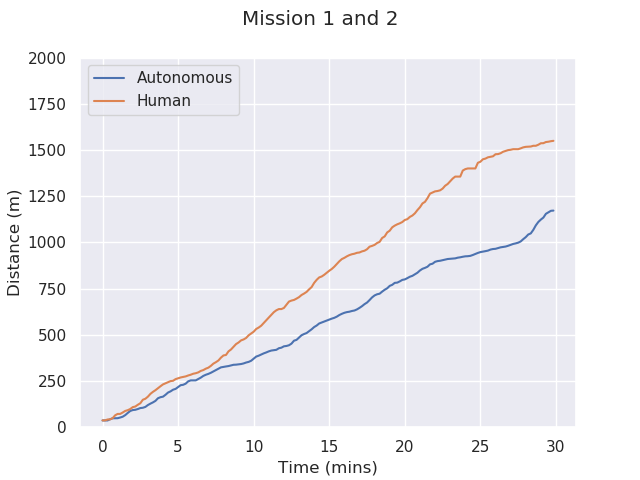}
  \caption{Course Performance for Mission 1 (Human-Robot Team) and Mission 2 (Robot Autonomy)}
  \label{fig:combined12}
  \includegraphics[width=0.45\columnwidth]{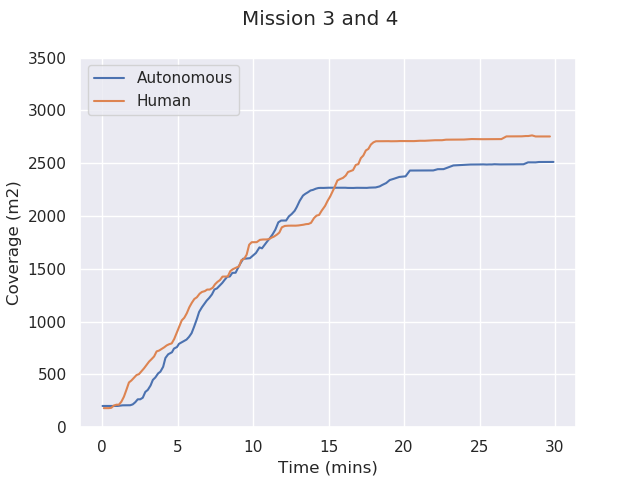}
  \includegraphics[width=0.45\columnwidth]{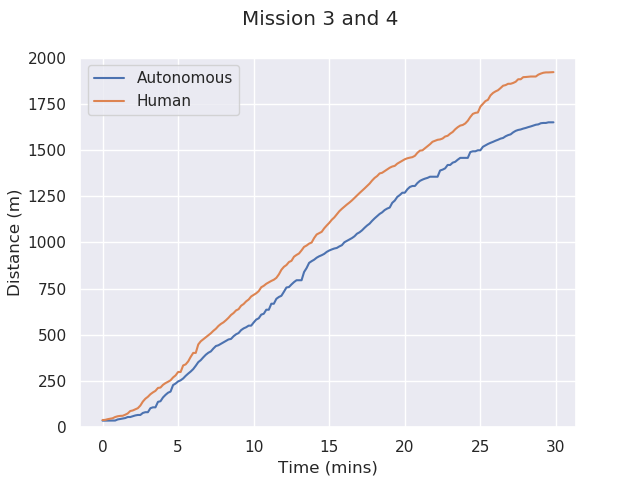}
  \caption{Course Performance for Mission 3 (Human-Robot Team) and Mission 4 (Robot Autonomy)}
  \label{fig:combined34}
\end{figure}

\begin{figure}[h]
  \includegraphics[width=0.45\columnwidth]{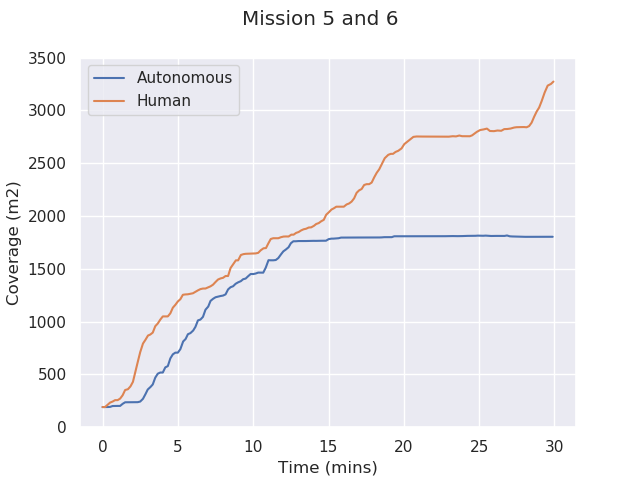}
  \includegraphics[width=0.45\columnwidth]{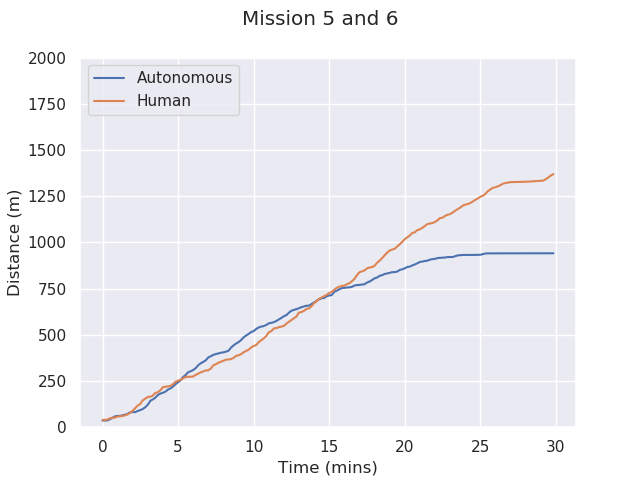}
  \caption{Course Performance for Mission 5 (Human-Robot Team) and Mission 6 (Robot Autonomy)}
  \label{fig:combined56}
  \includegraphics[width=0.45\columnwidth]{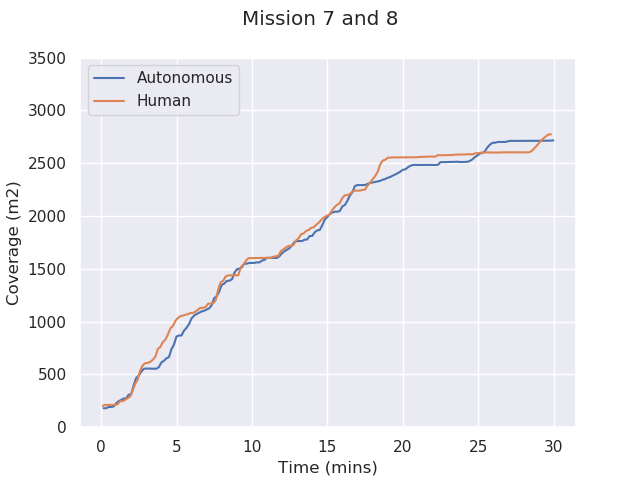}
  \includegraphics[width=0.45\columnwidth]{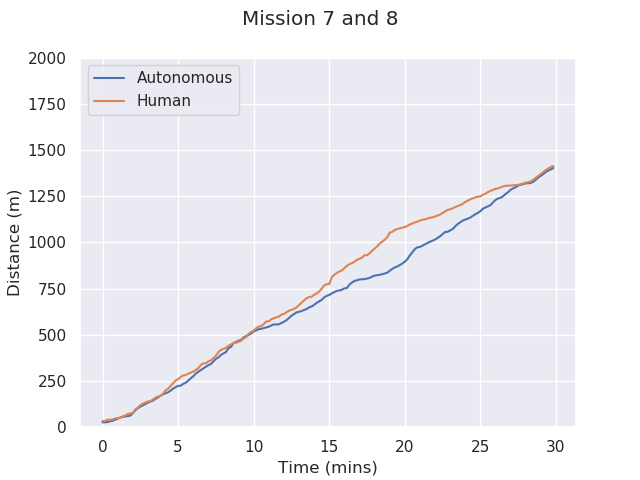}
  \caption{Course Performance for Mission 7 (Human-Robot Team) and Mission 8 (Robot Autonomy)}
  \label{fig:combined78}
\end{figure}

\begin{figure}[h]
  \includegraphics[width=0.45\columnwidth]{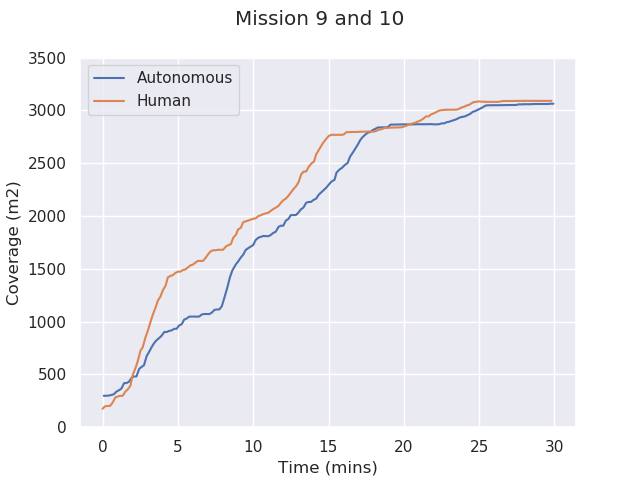}
  \includegraphics[width=0.45\columnwidth]{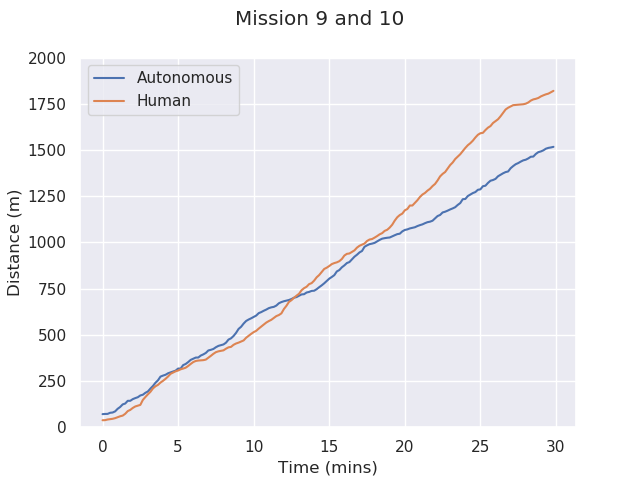}
  \caption{Course Performance for Mission 9 (Human-Robot Team) and Mission 10 (Robot Autonomy)}
  \includegraphics[width=0.45\columnwidth]{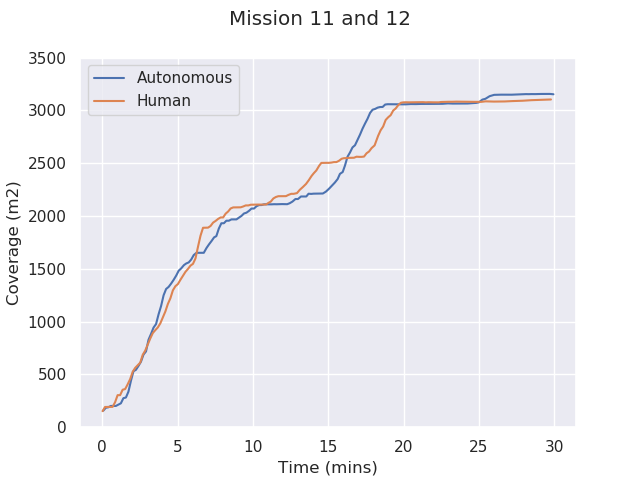}
  \includegraphics[width=0.45\columnwidth]{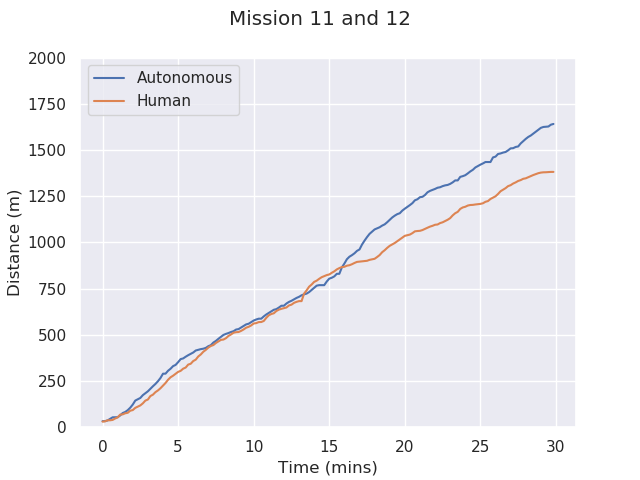}
  \caption{Course Performance for Mission 11 (Human-Robot Team) and Mission 12 (Robot Autonomy)}
  \label{fig:combined1012}
\end{figure}

As seen in Figure \ref{fig:combined12}, Mission 1 and 2 human operators had higher scores in both total coverage (left graph) and distance traveled (right graph), with Mission 1 and 2 seeing a converging point on total unique coverage but human operators having traversed even greater ground to achieve this goal. A similar pattern was seen in Figure \ref{fig:combined34} with robot autonomy course coverage at the 13\,min mark higher than the human-robot team coverage. Mission 1 and 2 used three robots whereas Mission 3 and 4 used four robots, but there were limited differences between the two matched-pairs for unique coverage, but higher scores for total distance covered. As seen in Figure \ref{fig:combined56}, the robot autonomy team was stuck in an area in which new ground was not easily covered by the robot, which was instead overcome by the equivalent human-robot team run through operator intervention. This was caused by addition of an obstacle which created a narrow constriction, making access to the section of the course beyond that point challenging. As seen in Figure \ref{fig:combined78} where the same course was used by the robot autonomy team, the final outcome was roughly similar in coverage and distance human-robot and autonomy runs. As seen in Figure \ref{fig:combined1012}, in Missions 9 to 12 all teams reached equivalent unique coverage with greater distance covered on one run for the autonomy condition, and another on the human condition. In both pairs, one team covered more ground than the other, but ended with equivalent coverage. Individual robot performance for total course map coverage and total distance can be seen in the Appendix. 

\subsubsection{Safety and Error-Related Events}
Safety and error-related events were measured based on the total number of times a robot was eStopped during the mission (See Fig ~\ref{estop}. Events related to eStop use included but were not limited to robots falling over or breaking down, robots coming into close proximity to people or buildings, fatal errors caused by the robots being stuck in difficult situations, and parts of the robot being trapped in certain areas. E-stop was applied under the following conditions: 1) until the first use of the robot in each run, 2) if a safety-related event occurred, in which case the robot remained stopped for the remainder of the run, 3) if a course-related event occurred, in which case the course issue was remediated, and then the robot was permitted to continue. For Mission 2, the robot was caught on equipment (R1) and a fall occurred that while the eStop was not applied, the incident was counted as a safety-related event (as denoted by the asterisk for R2/Mission 2 in Fig~\ref{fig:estop}). In Mission 3, the robot attempted to climb a steep ramp and the eStop was applied to preemptively prevent fall damage. In Mission 6, the robot pulled on temporary fencing, and in Mission 7, the robot fell into a ditch. One course-related event occurred with robot R5 in Mission 8 where the robot entered the staging area, which was intended to be closed off. The robot was temporarily eStopped, the course error remediated, and the robot was permitted to continue. As denoted by the asterisk in Fig~\ref{fig:estop}, this was not counted as a safety event.

Safety-related events included risk of damage to static equipment, or robots falling over. According to Figure \ref{fig:estop}, the Human-Robot Team condition had a lower eStop use rate (twice) compared to the Robot Autonomy condition (three times). These represent one stop per 563 min and 371 min of robot time, respectively, a 34\% reduction for the Human-Robot Team condition.

\section{Results - Qualitative Data}
The following section describes the operator interviews and analysis of their experience and expectations of the missions run, including discussion around notable events, challenges and key decision-making points. Six major themes arose from operator interviews: 1) perceived and actual need for human intervention, 2) specific scenarios that trigger operator intervention, 3) robot autonomy compared to operator choices, 4) robot failure, dangerous situations and events of error-related recovery, 5) operator cognitive load in response to challenging events, and 6) operation with more agents and team co-ordination. In summary, reasons for intervention included to speed up mission objectives, to cover more ground in an optimal way, to better control robots through rough terrain areas, to use higher-order knowledge to prioritise high-yield areas, and to maintain tighter coordination. After observing the autonomous missions, operators reported an increase in perceived robot competency, trust in robot teams to contribute to mission outcomes, and were less likely to intervene in future missions. However, operators also reported that autonomous robot behaviour was not always clear and understandable, even if autonomous robot teams achieved equivalent mission performance scores. Subsections and quotes will be presented in bold face below to highlight a brief summary of the theme and key takeaways. 

\subsection{Operator Interviews on Human-Robot Team and Robot Autonomy Runs}

\subsubsection{Perceived and Actual Need for Human Intervention}
Operators reported their attitude shift over time from close management to \textbf{\textit{learning to trust robot autonomy}}. This process was accelerated when operators watched autonomous runs on similar courses they had completed themselves. As described by one operator, ``I think this is a really good lesson, because the autonomy works really well, so the autonomy does split the agents up really well. It really doesn't take a lot of intervening by me'' [Mission 4, O1, CA] which helped to grow  ``confidence in the system'' [Mission 11, A, CH]. This method also produced a change in perspective on their future operation style: 

\begin{displayquote} ``I think it was a really good showcase today of what the autonomy can do, and even to the point where even the intervention I did have would have saved small amounts of time compared to what they did'' and that ``it just reinforces just leave them be, unless there's a clear-cut reason why you should intervene'' [Mission 8, O1, CA]. \end{displayquote}

This resulted in greater appreciation for robot autonomy, and learning to use the system at the expense of longer completion times: ``There were times when I could have - what I would have done would have saved time, for sure, but eventually, they got around to it before the end of the run, which was enough, but yeah. I definitely could have saved time, but they did make logical decisions'' [Mission 8, O1, CA] and that ``trusting in the autonomy but where you can, try and send multiple platforms'' [Mission 12, O1, CA]. Furthermore, while robot choices did not always appear to be transparent or logical from an external viewpoint, robot outcomes were at times, still beneficial: 

\begin{displayquote} ``I think I predicted incorrectly. I thought they chose to go in certain paths that would have led them to not having time to do the coverage that they did, but they very quickly did that coverage and surprised me'' ... ``Sometimes, it doesn't, but \textit{\textbf{it was very impressive to see it work in such a tight way today}}.'' [Mission 8, O1, CA]. \end{displayquote} 

\begin{displayquote} ``The big lesson from this for me was that the autonomous runs are actually very impressive, particularly on this course. The learning lesson, for me is to let the robots do their thing. The interventions that are required are really just if the robot’s doing something adverse, and that should be pretty obvious. Keeping an eye on it when it’s going through narrow gaps and taking over when necessary, but otherwise \textit{\textbf{trying to be hands-off}} for this course'' [Mission 11, O1, CH] \end{displayquote} 

\begin{displayquote} ``It \textit{\textbf{reinforces my confidence with the robots}} and the more I run with them, the more I see them run, the more confident I am and for a new course, you know, always going in with the mindset that you trust the autonomy before trying to take over is a very important thing'' [Mission 12, O1, CA] \end{displayquote} 

One operator reflected on a previous time in which intervention caused more harm than good, showing how operator involvement can instead be detrimental. One operator comments on a separate run that occurred outside of the experimental mission set in a cave scenario:

\begin{displayquote} ``There were moments that we [the operators] didn't have confidence in the autonomy. In the process of taking over from autonomy to manually intervened, we caused the robots to roll over and damage hardware, so the autonomy is improved to a level where the confidence in that autonomy is something that operators should trust'' [Cave Run, O1, CA].\end{displayquote}

When operators were asked about if they would change anything about their operation style, one operator said ``I think \textit{\textbf{I might rely on the autonomy a bit more}}'' and ``I think the autonomous runs have proven to me that they’re more capable than I gave them credit for. Just probably not as efficient as when I can jump in'' [Mission 10, A, CA], showing a growing relationship to robot autonomy trust to instead focus on other critical operator-related tasks. The same operator continued with ``there’s been some cases in the last three runs where \textbf{\textit{I was pleasantly surprised that the robots did as much as they did}}'' [Mission 10, A, CA]. As one operator reported, ``The very first time, it's a big shock, and how you operate on that course is very different to how you operate the last time'' [Mission 11, O1, CH]. The same operator also reported that \textit{\textbf{passively observing robot autonomy in action would be beneficial for inducting new operators}} to the system and to help build trust for new team members: 

\begin{displayquote} ``It's probably a good one for showing an operator what is possible if you just leave them do their thing, so that's an important thing. This is something we have struggled with over time, [Operator] and I, where you just want to take over because you don't trust them.  But we've been able to build that trust over time, and the capabilities of the autonomy has advanced significantly in the last six to 12\,months, which is amazing. So it's really good see it do this type of course, and if you were to train a new person, it would be great to let them know and experience what it's like for them to just do it'' [Mission 8, O1, CA] and ``If someone new came in, probably the first thing that they should do is just be really ingrained and confident in what the robots can do'' [Mission 12, O1, CA]. \end{displayquote} 

\subsubsection{Triggers for Operator Intervention}
Operators often reported different events and situations in which their involvement was considered to be advantageous or necessary to improve team performance or to meet mission objectives. Operator-driven interventions often involved adjusting the robots' movement to \textit{\textbf{reassign exploration points}} ``to guide the robot down a set path'' [Mission 7, O1, CH], to help the robots to \textbf{\textit{cover more ground}} to increase the opportunity to discover more artifacts, or to \textbf{\textit{better overcome challenging terrain}}. When operators did intervene, operators often stated that their involvement was minimal.  For example, ``I don’t think I had any issues with what the robots were doing'' [Mission 9, A, CH] and ``I gave it a couple of hints where to go, but I don't think they were necessarily necessary'' [Mission 1, A, CH]. Operator intervention was further described below in one mission run: 

\begin{displayquote} ``The robots were autonomous most of the time, very little teleoping, or waypoints or other manual interventions ... my involvement was very minimal'' [Mission 11, O1, CH].\end{displayquote}  

Operators were often content for the robot team to continue their objective: ``I think most of the robots were good in autonomy, and I helped out where I could. As far as I know'' [Mission 9, A, CH]. Operators were also reporting that they would be \textit{\textbf{comfortable to leave the robot team alone when the team was performing well}}. For example, ``I wouldn't have had done anything. They seemed to be going perfectly'' [Mission 2, A, CA]. At times, operators reported an intervention need to \textit{\textbf{take a more direct approach to rapidly address priority areas}}. For example, ``I don’t think I needed probably any involvement to be honest, but I think I helped speed up some sections'' [Mission 9, A, CH]. Involvement was also considered more \textbf{\textit{necessary for events or scenarios that were time-sensitive to complete}}. For example, ``I did guide them [the robots] down the back of S block, and that was just a case of saving time'' towards the end of a mission run. There were many other teleoperation events reported to save time and redirect to the mission outcome with another mentioned below:

\begin{displayquote} ``There were a couple of cases where robots were being a bit slow, because they were caught up in nearby obstacles, I would give them a quick teleop touch out of the way. Other than the teleoperation and the prioritisation regions, I didn’t have a lot of input in the robot navigation'' [Mission 9, A, CH]. \end{displayquote} 

Furthermore, ``they [the robots] would need to burn through all of their exploration points before they'd potentially cover that area'' [Mission 3, O1, CH]. Another example is described below: 

\begin{displayquote} ``Planning that has to happen for the robot to run autonomously takes time. It takes time to plan. The max velocity off the robot is linked to how far away it is from its path, and I can bypass both of those things using my human brain and go max velocity, which is something that the autonomy can't do easily, or at this level that we have it'' [Mission 7, O1, CH]. \end{displayquote} 

Operator intervention was also perceived to be required to help robots to \textbf{\textit{navigate challenging terrain}}:

\begin{displayquote} ``Some of the ATRs [robots] required a bit more hand-holding. That might be because the course was slightly more difficult in terms of some of the dead ends that it had and having to turn around and some constrictions that were added'' [Mission 5, A, CH] \end{displayquote} 

This intervention type was also used \textit{\textbf{in areas with greater constrictions or dead-ends}} compared to open-spaced areas. For example, ``the difficulty may have been the addition of the playground (See Fig ~\ref{fig:terrainchallenges}. Second image from the right on the top row), but I pretty quickly cleared that out. It didn’t seem to be as maze-like as the last one, there seemed to be fewer constrictions'' [Mission 9, A, CH]. Other reasons included to help find artifacts when the robots had not found one yet, or were perceived to be less likely to find it without operator support. In an autonomous run, ``one artifact that wasn't detected was visible in another frame, so I possibly could have seen that where it wasn't detected by our autonomy'' [Mission 4, O1, CA]. This was further explained below:

\begin{displayquote} ``There were two objects which I had to MID (a map-informed detection, which means the detection that we actually got didn't seem to be perfectly correct, so I had to click around it to find the object). I did that for at least one object, so I was directly involved with one scoring point which I don't think would have been possible without an operator'' [Mission 1, A, CH] \end{displayquote} 

Operators also believed that they were \textbf{\textit{better positioned to reason about when an area was considered to be complete}}, and requested actions because of it. For example, ``There was no obvious areas of the course that I could see that I had missed'' [Mission 6, A, CH]. Other example actions included to improve operation success by taking into consideration other events that might happen during the run, such as ``to split the robots up to have redundancy for the other agent that went into the dark by itself'' [Mission 4, O1, CA]. 

Operators also had different styles relative to their preferred control level over the robot team. This included \textbf{\textit{managing time and resource intensity of direct operation through teleoperation at the expense of task awareness}}, stating that its ``not a good idea when you've got multiple robots running around''. The last resort was identified using ``teleop when the robot can't do something that you need it to do.'' [Mission 7, O1, CH]. Operators also acknowledged that their \textit{\textbf{operation actions changed based on key information and constraints}}. In one instance, teleoperation was justified by the following event: ``I could see obviously the Spot [robot] went down and the ATR [robot] wasn't making fast enough progress, and I know the time was running out. It's always faster to get on the sticks if I've got good comms, and it's much faster to get the robot to where you want it'' [Mission 7, O1, CH].  When asked about operator decision-making for switching from autonomy to teleoperation, one operator reported ``\textit{\textbf{panic}}'' and \textit{\textbf{knowing ``that there are things left in the mission}}'' as well as when ``it's fairly \textit{\textbf{obvious when a robot can't do certain things}}'' [Mission 7, O1, CH]. Two examples are provided below: 

\begin{displayquote} ``Although I only spotted it [an open door] at the last second, I did prioritise getting that robot to that location and probably spent the last five minutes trying to get that robot in there and as far along that section as possible.'' [Mission 6, A, CH] \end{displayquote}

\begin{displayquote} ``It did a reasonably good job of autonomously getting to the waypoint that I set, but at some point, I decided that it wasn't going quick enough, so I took over, grabbed the controller and teleoped it as fast as I could into the side of the branch. Unfortunately, wasn't able to complete the entirety of that section of the course, so I probably left some stuff on the table, and that might be where the two objects that I couldn't find were located'' [Mission 5, A, CH] \end{displayquote}

Reasons for intervention also included the belief that the robot was not capable of completing the task, especially ``if there's no other way to do it'' [Mission 7, O1, CH]. Other intervention factors that influenced their need to take control included \textit{\textbf{operator mood}}. For example: 

\begin{displayquote} ``I'm in a bit worse of a mood than I was yesterday, just because of robot troubles. I definitely felt the need to or the want to intervene a bit more often. Whether that's because the course was more difficult or because I just was in that sort of temperament'' [Mission 5, O2, CH]  \end{displayquote} 

Considering the robot autonomy runs, operators clearly noted specific events or situations in which \textbf{\textit{if they were able to intervene, they would have taken different actions}}. For example, ``I could see the two robots were converging on the area, and they got distracted by some nooks and crannies. I could have just expedited that'' [Mission 8, O1, CA]. Another example was as follows: ``I would say as - if as an operator I could have jumped in, I would have scored two extra points, which would have ended up beating the first run, I believe'' [Mission 2, O2, CA]. A further example is as follows: 

\begin{displayquote} ``There was definitely a couple of cases where I would have intervened but didn't, and whether that helped or not, I'm not fully sure. But, for example, I saw the Spot go into the grass area, and as an operator, I would have intervened and grabbed it and moved it away, and it potentially would have stopped it from falling.'' [Mission 2, O2, CA] \end{displayquote} 

Operators also had the important \textit{\textbf{advantage of foresight}}, as well as a \textit{\textbf{more rigorous understanding of future events and situations}} in which each robot could have been used to the best of its mechanical ability to explore course sections. 

\begin{displayquote} ``The only thing I would have done with the Spot [robot], as well, is I would have sent it into the other section of the course, because from what I had already seen of the course from the first agent, there seemed to be more areas for the Spot to explore'' ... ``I would have put Spot into the tunnel rather than in the direction that it did go.'' ... ``because in the tunnel, there's more nooks and crannies and diverging branches and things that I think a Spot [robot] is better at doing.'' [Mission 2, O2, CA] \end{displayquote}

\begin{displayquote} ``Somewhere towards the end of the run, I realised that there was a third branch, probably a large part of the course that I had missed completely, and so at that point, I tried to get bear out of the tunnel as quickly as possible to get over there'' [Mission 5, O2, CH] \end{displayquote} 

Operators also reported the \textit{\textbf{importance of information collection to make their next decision and how to best intervene}}. For instance, ``when the robots have a motor fault, I don’t really have a way of knowing, I have to wait until I get to that robot and see that it hasn’t moved'' [Mission 11, O1, CH]. In addition, ``information forwarded up to me, and more of an alert system, would be beneficial to get to the robots more timely'' [Mission 11, O1, CH]. Operators often acknowledged the \textit{\textbf{importance of knowledge discovery in their operation method}}, stating that when learning about important information or knowledge about the situation, different choices could have been more effective: ``Ideally, I would have spotted that third section of the course earlier, and I would have taken that third robot and put it into that section first'' [Mission 5, O2, CH]. Another example is as follows: 

\begin{displayquote} ``In the tunnel, there is a section leading from the outside the tunnel area back into the tunnel that has a bit of a zig-zag, narrow corridor, and as an operator, I would have just put a waypoint at the end of that, so that it could get through that constriction non-autonomously - well, semi-autonomously, but it needed that little bit of a push from an operator'' [Mission 6, O2, CA]  \end{displayquote} 

\begin{displayquote} ``There was an area in the tunnel that was not explored at all during the run and it would have been great if the ATR had sort of figured that out and gone there but it didn’t happen'' ... ``I definitely would have wrangled the ATR to explore the sections of the course that we missed'' [Mission 10, O2, CA] \end{displayquote} 

Future operator features were also mentioned, bringing attention to new methods that operators could use to better improve their performance, and to reduce the need to intervene, such as ``colourising the point clouds'' to create information that would be ``very valuable for an operator'' to use [Mission 10, O2, CA]. 

\subsubsection{Robot Autonomy  Compared to Operator Choices}
Operators used the opportunity to view the robot autonomy runs as a learning experience, and to compare actions and outcomes across both conditions. For example, ``The ATR [robot] that went up the tunnel was slower, in general. I think in the first run, we got to the end of the tunnel earlier than we did last time'' [Mission 2, O2, CA]. In this process, \textit{\textbf{operators often reported varying opinions between their interpretation of robot autonomy choices}}. After some runs, operators reported being impressed with robot autonomy choices and options: ``It’s really good to see how the robots split themselves up'' [Mission 12, O1, CA]. In other course runs, robot autonomy appeared to be strongly suited to the course layout, which \textbf{\textit{in some cases, the robot team surprised the supervising operator on its capacity to independently complete the course}}: ``they [the robots] really didn’t waste a lot of time. There wasn’t even a whole lot of opportunities where I would have been faster if I directed them. They really smashed it out.'' [Mission 12, O1, CA]. In some course runs, operators were also viewing robot autonomy behaviour that they believed could have outperformed their own run: ``\textbf{\textit{I think the robots actually covered the course quicker than I did, if I’m being honest.}} They seemed to get the detections before I did''. [Mission 12, O1, CA]. 

Operators still encountered some surprising outcomes: ``there were a few reports that I didn’t report until very close, or at the end of the run, due to being \textit{\textbf{distracted by robots doing cool things}}'' [Mission 12, O1, CA]. The most evident was that robot autonomy choices were sometimes perceived by the operators to be less efficient and direct in their efforts to reach the mission outcome of finding artifacts. For example when robots were searching for items, ``some of those robots weren't really all that productive'' [Mission 4, O1, CA]. In some cases, operators reported their acknowledged risk/reward trade-off when they were reflecting on choices about when and how they would have intervened if they were operating the team while observing the robot autonomy course run. For example: 

\begin{displayquote} ``It's slightly different to what I would have done, because usually, like I did in the last run, you send in a second robot as a redundancy measure, but one Spot was able to clear that whole tunnel by itself, which was very good. But also, if something had have happened, we wouldn't have been able to do anything about it'' [Mission 4, O1, CA] \end{displayquote}

\begin{displayquote} ``In the first run [Mission 1], Spot explored underneath the landing in the tunnel, whereas in this run [Mission 2] it didn't, because the Spot went in a different direction. The ATR that did go into the tunnel went all the way to the end of the tunnel and then didn't have time to come back and look at all the nooks and crannies'' [Mission 2, O2, CA] \end{displayquote}

\begin{displayquote} ``It’s a lesson for us to try not to touch them where we can and it’s really - as I’ve mentioned over the runs, it’s only if they choose an incorrect path or it’s maybe to get a bit of speed that the operator could intervene and guide it down a path. However, in this case today, they did exactly what I would have done'' [Mission 12, O1, CA].\end{displayquote}

Operators also had extensive experience with the human-robot team setup, which suggests that \textit{\textbf{operators had already come across most scenarios in the past}}. For example, ``I feel like I've seen it all at this point.'' [Mission 2, O2, CA], ``often you do encounter narrow passages and you've got to deal with those, but they're not wholeheartedly unexpected'' [Mission 7, O1, CH], and ``it doesn’t necessarily surprise me that it chose to go in the direction that it did. \textit{\textbf{Sometimes, it's a 50/50 call, and it chose the direction that it chose}}'' [Mission 2, O2, CA]. However, viewing robot behaviour without being able to intervene was reported as ``a bit frustrating to see the robots do the wrong thing'' [Mission 10, O2, CA]. This also involved \textit{\textbf{operators reporting that the robot team did not always adequately cover certain areas}}. For example, ``The only part that I had to take a second look at, there was a very narrow opening at the barrel area (See Fig ~\ref{fig:humanchallenges}), I just wanted to make sure that I covered it, so I sent a robot back there once'' [Mission 11, O1, CH]. Furthermore, certain course orientations did create some some unusual circumstances for robots to behave in unexpected ways. For example: 

\begin{displayquote} ``It's a little bit unexpected that the robots couldn't get past - in the S block constriction (See Fig ~\ref{fig:humanchallenges}. Bottom right image), the robot did get past the constriction but then didn't continue on. That's a little bit surprising that'' ... ``a task wasn't generated at that point for it to continue'' [Mission 6, O2, CH] \end{displayquote}

\begin{displayquote} ``I think the other robots had sort of taken up all the big frontiers already and so it was just trying to find all the nooks and crannies and other frontiers and tasks that it could do and didn’t have the awareness to realise that there was a chunk of the course that it should have gone towards.'' [Mission 10, O2, CA] \end{displayquote}

Operators also drew attention to inconsistencies in which the robot autonomy made decisions around its organisation and \textit{\textbf{exploration choices that would have been different if the operator was involved}}: 

\begin{displayquote} ``Not nearly as good as the first run, purely because there were some hard and difficult constrictions to navigate, which really seemed to require operator input. Similarly, with the constriction with the chair behind S block, it just needed an operator to teleop it past the chair and a bit further beyond the chair, so it started generating tasks again, or else a couple of waypoints probably would have done the trick as well'' [Mission 6, O2, CA] \end{displayquote}  

\begin{displayquote} ``The \textbf{\textit{only difference would have been just some very high-level guiding.}} There isn't really anything else I would have done to change what they did, so I don't think they made too many errors with regards to where they went'' ... ``the only thing I could have done is to direct them at a high level, but eventually they got there anyway'' [Mission 8, O1, CA] \end{displayquote}  

At times, viewing robot autonomy without being permitted to intervene evoked a notable emotional response from the operator, showing insight into how the operator processed the concept of an autonomous robot team that could score higher without an operator involved at all. Over time, operators were starting to be drawn to the notion that their assistance may not be needed at all, challenging the commonly held idea that operators are used to closely direct and control autonomous robot teams to achieve better mission outcomes, indicating a potential phase shift from ``operator as necessary'' to ``operator as optional'': ``[there] seemed to be the right number of robots and they selected the directions to go very logically. There weren't any areas that were missed and it was very similar to how I ran, actually'' [Mission 12, O1, CA]. 

Despite robots having a strong level of intelligence and autonomy to complete the mission on their own, there were several times in which operators were \textbf{\textit{unable to decipher their intention, leading into limited levels of understanding and explainability in the robots behaviour}}. For instance: ``operation outdoors in general is not exactly the way the autonomy is programmed to operate so it does get stuck sometimes in outdoor environments because it sees gigantic frontiers that it thinks it should do.'' ... ``So it’d be ideal if it could understand that no, you’re good, you’ve explored this area, go do something else'' [Mission 10, O2, CA]. When operators were asked about ways the robots could behave that would improve their level of use and explainability to the operator, a few suggestions were provided: ``better tuning in situational awareness for robots to avoid difficult areas'' and ``in terms of autonomy, being able to handle outdoor environments with huge frontiers is probably one of them if exploration in outdoor environments is something that we’d want to tackle'' [Mission 10, O2, CA]. In addition, ``I think reliability of this system in general, I think I’ve realised is still not there and that does increase the stress of the operator when - especially if the operator’s involved in getting the robots up and running'' [Mission 10, O2, CA]. Furthermore: 

\begin{displayquote} ``It’s very possible that if I’d started with Spot, it could have gone into the playground and it almost certainly would have catastrophically fallen over. \textbf{\textit{Luckily, autonomy chose not to do that but that would be something that I would increase my belief in the robot’s capabilities }}if I knew that a Spot could sense that hey, this area is dangerous, I’m not going to go in there'' [Mission 10, O2, CA]. \end{displayquote}

\subsubsection{Robot Failure, Dangerous Situations and Events of Error-related Recovery} 
Across all missions, \textbf{\textit{robot failures and error-related events occurred in both conditions, even when operators were involved}}. More often than not, these \textbf{\textit{events were often low risk/damage}}, such as the robot ending up in scenario or situation that was not favourable. This included the robot being \textbf{\textit{stuck in dead-end corridors or being unable to navigate its way out,}} which often did not result in critical outcomes. In one example, ``there was one time when a Spot robot walked on another Spot robot and had to be eStopped for safety reasons and restarted'' [Mission 8, O1, CA]. Operators \textbf{\textit{had some necessary tools to mitigate some low-level errors and failures well ahead of time}}, and that ``it was very good to see the controllers work as they've been designed and getting into those tight spaces'' [Mission 8, O1, CA]. Certain error types also required direct operator involvement to successfully overcome the event with one example presented below:  

\begin{displayquote} ``I think one extra entrance was blocked, and there were a few other little nice things that would have blocked the ATRs [robots], but they managed to push their way through autonomously in one case, and in one case, it was a lot of manual intervention'' [Mission 7, O1, CH] \end{displayquote} 

In other runs, the robots ended up in more serious or hazardous places. For example, ``I saw Spot [robot] in dangerous positions'' [Mission 10, O2, CA] and in an autonomous robot run, the robot was hooked on an \textbf{\textit{equipment piece}} [Mission 2, O2, CA] that required the immediate use of the eStop. \textit{\textbf{Failures and errors were also attributable to support systems that were in operation }}during each run. One operator described this as follows:

\begin{displayquote} ``I think the other part of that is the comms situation, which is probably a different side to the story.  But often, that is a big component as to why you can't go somewhere, why you don't get data back or why your planning doesn't work, if a node doesn't come online'' [Mission 3, O1, CH] \end{displayquote}

Further examples are provided below: 

\begin{displayquote} ``I think it got too close to some of the orange netting fencing and may have even started to chew it up into the tracks, at which point the safety operator e-stopped it, and I think that was eventually pulled out of the tracks right at the end of the run'' ... ``Similarly, an ATR came back to the pit area, I think it got too close to some of our charging equipment, and the safety operator briefly eStopped it'' [Mission 6, O2, CA]  \end{displayquote}  

Some \textit{\textbf{possible error-related events were often prevented beforehand}}, such as being cautious ``when robots start up together'' [Mission 12, O1, CA]. Operators also reported their capability level to prevent or correct failure, errors or dangerous situations. For instance, during one robot autonomy run, ``Spot did get into a situation that it shouldn't have, but again, that's one of those cases that an operator would jump in and try and resolve that issue'' [Mission 2, O2, CA]. In response to failures and to mitigate damage, \textbf{\textit{operators helped to nudge the robots to a path or goal that resulted in less risk}}: 

\begin{displayquote} ``It was just simply picking a path for them to go, and then the parts at the end where they were getting quite close, it was just mitigating damage and making sure that they separated, went one way, and the other went the other way'' [Mission 3, O1, CH] \end{displayquote} 

\begin{displayquote} ``If it’s a fully autonomous run, Spot [robot] would need to be able to sense the environment better and know that, for example up here, there’s big rocks with big gaps in between. The resolution of all of that [unclear] just isn’t enough for it to realise that that’s dangerous and it might get a foot stuck in there and fall. In terms of an operator run, because I have some amount of prior knowledge, I know that that’s dangerous but if I could get, again, better - finer resolution of the course terrain, I could sort of be able to tell the robot if it couldn't figure it out itself, that it’s more dangerous than it knows'' [Mission 10, O2, CA] \end{displayquote} 

Operator involvement and absence from involvement had varying results during dynamic situations. For instance, ``\textit{\textbf{one of the ATRs [robots] did collide with an obstacle in a difficult location}} that we are aware of being difficult. I don't think that's anything to do with an operator versus autonomy. It's \textbf{\textit{just the luck of the draw sometimes}}'' [Mission 2, O2, CA]. Furthermore, one operator stated that it was \textit{\textbf{possible that the robots did enter dangerous positions or areas, but they were simply not aware of the situation}} occurring from their operation viewpoint. For instance, ''as far as I know, they didn't get into any precarious or dangerous positions, there was no e-stops, no catastrophic failures'' [Mission 9, O2, CH]. Operators also acknowledged their limitations to fully protect robots from damage: ``there are definitely environments out there that I have no knowledge of that if I was to send a robot in, I would not have adequate situation awareness to protect the robot from doing dangerous things'' [Mission 10, O2, CA]. This was followed by this potential suggestion:

\begin{displayquote} ``If I could get that feedback in some form to show me that this area is potentially dangerous for a robot, that would help me as an operator protect the robot from doing things that are potentially dangerous for itself and others'' [Mission 10, O2, CA] \end{displayquote} 

Other times, mission outcomes were impacted by chance factors such as \textbf{\textit{environmental terrain that can cause the robot to fall over}}: ``It could have fallen over at the start. Sometimes, they fall over. Once they fall over, we don't have a way of correcting and re-standing at this point'' [Mission 7, O1, CH].   

Operators also reported \textbf{\textit{allowing the robot team to attempt some difficult scenarios on their own prior to intervening in the situation}} to assist with rerouting the robot back on track. In one example, ``it was a very tight door, very tight scenario, and potentially we could have got in there autonomously, but again, it would have taken a lot of time, and the robot had already tried a few times'' [Mission 7, O1, CH]. Other types of safety-related events were higher-risk than others, but nearly all of these events occurred during the robot autonomy condition because operators were not able to intervene to either prevent the failure or event from being likely to happen in the first place, or to take rapid actions to mitigate the impact of it.

The missions also provided new insights into how operators could \textit{\textbf{improve their operation style to avoid future setbacks and error-related events}}. Operators also acknowledged that their involvement in the run may have also had adverse effects, stating that in some instances, it is more beneficial to leave it to the autonomy to resolve its own issue instead of intervening at the risk of greater damage: ``I probably would have done more damage had I tried to take over from the door controller, which I had to do in my run, because it was wasting a lot of time'' [Mission 8, O1, CA]. This included \textit{\textbf{building in more fail-safe behaviours}}: ``what I probably should have done, if I did this run again, was always follow a Spot with another robot. That's a pretty good lesson from this run'' [Mission 7, O1, CH]. This included for recommending new features that could also contribute to reducing cognitive load, so that ``someone else could come in and be able to operate much easier without having to button hop and grab different controls'' [Mission 12, O1, CA].

\textbf{\textit{Some level of responsibility for robot failure was reported by operators based on course design rather than robot performance}}. In commenting on one error-related event, ``we left a rogue Spot [robot] in a traversable area, so it certainly wasn't the fault of the robot, and the robot recovered well'' [Mission 8, O1, CA]. In relation to recovery from error, there were some instances in which errors provided limited impact to the overall run and a simple restart was all that was needed to continue the robot to the mission goals. Given the experimental set-up, it was also acknowledged that \textit{\textbf{should a robot have had a similar error in other real-world scenarios, robot recovery would not have been possible}}, which is likely to cause greater impact to mission success. Furthermore, \textbf{\textit{error-related events prompted operators to use new strategies in the remaining missions to mitigate their effects}}: ``I didn't at this time, send both ATRs far away from the Spot. One ATR was close to the Spot at all times, which I didn't need, but it was a good peace of mind'' [Mission 11, O1, CH]. 

\subsubsection{Operator Cognitive Load in Response to Challenging Events} 
In response to challenging events, experiment operators were highly experienced, which resulted in strong performance during most of the course run. For instance, operators reported that there were \textit{\textbf{few events or scenarios that were surprising}}, or events that had not seen before in the past: ``It wasn't really anything I haven't seen before'' [Mission 7, O1, CH] and ``I think \textit{\textbf{everything they did was very logical}}'' [Mission 8, O1, CA], but at the same time, ``always expect the unexpected, so \textbf{\textit{some things went wrong, but nothing too crazy}}'' [Mission 1, O2, CH]. Unexpected errors or challenges in communication methods between the operator and robots also contributed to a large portion of reported cognitive load, especially \textbf{\textit{when events did not go according to the operators level of expectation}}: 

\begin{displayquote} ``One of the objects seemed to have potentially been incorrectly placed on the scoring server, because I found it and I localised it well, as far as I could tell, and sent a bunch of reports'' [Mission 1, O2, CH] \end{displayquote} 

Stress was also a common factor that was reported to affect cognitive load: ``\textit{\textbf{If robots aren’t playing nice, the stress levels of the operator increases just in general, which leads - I think leads to poorer performance}}'' [Mission 10, O2, CA]. The same operator reported that they achieve better performance ``as an operator when I’m calm at the start and if my stress levels are increased by external factors prior to the run, that’ll carry over into the run and I think degrade my performance as an operator'' [Mission 10, O2, CA] and that the robot, course or other external factors can ``slow everything down, which sort of builds the stress levels because obviously we’re on a bit of a time schedule'' [Mission 10, O2, CA]. Furthermore, ``\textit{\textbf{shorter runs definitely increased the stress levels}}'' [Mission 9, O2, CH]. Cognitive load was also unintentionally attributed to the experiment itself: ``as part of the team that’s setting up these experiments to some degree, that stress level can carry over occasionally and that, I think, effects my ability to operate, to some degree'' [Mission 10, O2, CA].

Experimental runs presented unique events or scenarios, which required operators to become more involved and therefore, to exert more effort to overcome these obstacles. Operators reported that certain \textit{\textbf{direct hands-on functions were high on cognitive load, reducing the potential to conduct other critical tasks}}. In the instance of teleoperation, ``\textbf{\textit{as soon as you have to grab the joystick, you can't do anything else}}. If there is something else that needs attention, that is in your back of mind'' [Mission 7, O1, CH]. This can push operators into making critical decisions based on mission outcome gains. For instance: ``if one ATR can get through a door, down the back of W block, and the other ATR needed to be teleoped to the top of the hill, I couldn't do it. I would have had to have chosen one''. [Mission 7, O1, CH]. The operator further commented using the following scenario as an example:

\begin{displayquote} ``There'll be an indication of which one you can switch back to autonomy. So if you can give a robot 10 seconds to get it through something and then switch it back to autonomy, obviously that's a priority, but yeah, it really depends. It's not an easy thing sometimes to change a strategy partway through a run and decide what's the best thing to do, given the time that you have. Yeah, so intuition is a hard thing to know'' [Mission 7, O1, CH]. \end{displayquote} 

Operators commented on the cognitive load difference between the same runs when comparing to operator involvement compared to robot autonomy alone, which included the cognitive process to self-regulate their perceived need to intervene: 

\begin{displayquote} ``It's probably less stressful in the second run [robot autonomy]. It's a little bit more frustrating, if you make the distinction between stress and frustrating, because I wanted to jump in. I wanted to help the robots, but it was definitely less stressful in the second run, because I could just sit back and watch robots and don't have to worry about making bad decisions'' [Mission 2, O2, CA] \end{displayquote}

Reported cognitive load and operator stress was also mitigated to some extent in the course runs given that other trained safety operators were monitoring robots during the course. Therefore, distributed responsibility for the operation of robot teams can have an important influence on operator cognitive load and their direct contribution to risk or mission failure: 

\begin{displayquote} ``I know there's safety operators out there, so I can offload that worry to them.'' ... ``If it was just me monitoring the robots, I might have been more stressed, because I would be responsible for stopping robots if they're in dangerous situations.'' ... ``I really don't concentrate too much on the safety aspects of the robots, because I know there's people watching them'' [Mission 2, O2, CA] \end{displayquote}

\subsubsection{Operation with More Agents and Team Co-ordination}
This experimental set-up required operators to control, monitor and direct a heterogeneous multi-agent team of 3 to 4 robots over a course with many obstacles and roadblocks, which had its own advantages and disadvantages. First of all, more robots can be more strenuous, given that ``working with three agents in such a short amount of time maybe increased the skill level a little bit'' [Mission 9, O2, CH]. Furthermore, ``having three agents means you can watch them a bit better, there’s less load on the operator to manage them, whereas previously, when you run with four, five or six agents, you literally \textit{\textbf{cannot micromanage too many agents at all}}'' [Mission 9, O2, CH]. In the course run, when allocating different robots in different areas, one operator said that ``it reinforces the point that it’s \textbf{\textit{better to have }}multiple robots covering - \textit{\textbf{multiple different morphology robots covering the same area}}'' [Mission 12, O2, CA]. However, the success of these arrangements can also be by chance as well: ``it’s a lot of luck that the Spot [robot] went that way in the autonomous run but this is a case in point where you want robots with multiple morphologies kind of in all areas'' [Mission 12, O2, CA]. In addition, operators also noted an ideal ratio of robots to mission objectives and course size: ``I think in this size course, more robots wouldn’t help but obviously scaling the course and having more branches is ideal to have more robots to do that'' [Mission 12, O1, CA]. One operator also acknowledged that its ``more efficient for me to just be doing the high level direction and going through some of the reports'' than team co-ordination [Mission 12, O1, CA], but that \textbf{\textit{more robots could result in reduced performance}}. This was further explained below: 

\begin{displayquote} ``I think in this course size with more robots, it would possibly complicate things. We've seen in previous test cases where robots do interfere with each other but this is a good example of where robots can go and do their thing. So timeframe, if there was a longer run, it’s pretty impressive what we can do in half an hour but an hour run, it doesn't detract - having done hour runs before, the robots behave very, very well over that time. They’re very robust platforms'' [Mission 12, O1, CA]. \end{displayquote}

In completing the mission, there were some notable decision-making options for how operators could organise their team and how robots chose to organise themselves, such as using a consistent \textbf{\textit{strategy each time to achieve high success levels based on the robots strengths}}. For instance, ``Spots do very well in the tunnel, so I've kept that strategy this whole time'' [Mission 11, O1, CH]. \textbf{\textit{Mission outcomes influence choices for team arrangement }}as one operator stated that ``it does influence the result, depending on the type of platform and what direction they take'' [Mission 8, O1, CA]. Operators reported that there is \textit{\textbf{an optimal robot-to-course ratio that can provide the most benefit at the expense of operator involvement}}, but more agents can also be more challenging. For instance, ``this time, having three agents was more difficult, because the course was more difficult.'' [Mission 5, O2, CH]. When asked about team co-ordination, one operator reported that ``\textit{\textbf{the hardest decision is determining where Spots versus ATRs should go}}'' but that after these decisions were made that ``everything else was pretty standard in terms of decision-making'' [Mission 9, O2, CH]. The \textit{\textbf{robot team formation and number of robots also influenced how operators make decisions on assigning areas to certain robots}}. One operator described their decision-making points around team co-ordination below: 

\begin{displayquote} ``With three robots, it’s a bit more difficult to know which way to send two robots, so initially I sent two robots down to the barrels and behind S-Block, and half way through that, I switched that and sent one of those robots to go back to the tunnel. I’m hoping that was an important decision, because I feel like we may not have explored part of the tunnel if I hadn't done that'' [Mission 9, O2, CH] \end{displayquote} 

Operator decisions to \textbf{\textit{arrange heterogeneous multi-agent teams to ensure robot safety}} was also a key decision point. For instance, ``I could tell that the playground area of the course was dangerous for robots, so I made sure that no other robots would go in there once it was satisfactorily explored'' [Mission 9, O2, CH]. One operator described their use of the robot team based on environmental constraints and mission objectives, for instance, ``Spots are very capable and they're very fast, but obviously, ATRs can do this course very well'' [Mission 8, O1, CA]. One operator also commented on the \textit{\textbf{balance between mission objectives, number of robots, and time limit}}: ``I’d probably have said four would be the limit, in a half hour. The course would have to probably get substantially bigger to warrant any more than that'' [Mission 9, O2, CH]. Lastly, in some runs where the operator was using three robots, having an additional robot could have been beneficial to ensure greater confidence to explore all possible areas: 

\begin{displayquote} ``I would hope that if we had that fourth agent, I could have sent it to explore more thoroughly in the right side of the course, and I think potentially we might have found that last object that we were missing. You never really know, but I would hope that that would be the case, because I have a feeling that we might have just sent - the only agent we sent down the right side of the course might have skipped past an area and not thoroughly explored enough to find that object''[Mission 1, O2, CH] \end{displayquote} 

\section{Discussion}
The presented experiments explore how operators contribute to robot team performance.  Experimental results found that there were notable differences between human-robot teams and full robot autonomy on key metrics such as mission score, time to first artifact discovery, total number of eStop use, total distance and unique coverage.  Human-robot teams with operation led by a trained operator were more likely to cover unique ground in a shorter period of time, travel a longer overall distance with the robot team and have fewer events related to eStop usage, but required increased perceived operator effort to manage the operation. One matched pair mission was equivalent in performance when the terrain was more traversable and the course layout was more predictable, demonstrating that in favourable conditions, robot autonomy can perform as well as human-robot teams. The operator interviews provided further explanation and understanding behind operator actions, their reported level of involvement and direct contribution to human-robot team performance. 

\subsection{Overcoming Challenges and Reasons for Intervention}
Human operator contributions were valuable to overcome major issues and failure-related events. Operator involvement appeared to be most critical during events that could have caused damage or harm to the robot, during time sensitive events, or to help the robot from getting stuck in challenging areas that would prevent further progression. Operator intervention in these event types allowed autonomous robot teams to continue to maintain steady progress, while robot autonomy teams were at times stuck, or failed to break through a major challenge in the course run. Operators often intervened in an attempt to speed up mission run time to take shorter paths to cover new ground, to better teleoperate robots through tough terrain, to use higher-order knowledge to prioritise areas that were more likely to have artifacts, and to have an overall tighter control over robot operation and coordination across multiple robots. While operators do improve team performance scores, they can introduce delays in team coordination, which could be critical depending on how urgent or serious a 30-70\,s delay to select the next robot action would represent for the mission. 

Operators reported that their direct involvement contributed to a greater sense of mental demand, physical demand, temporal demand, level of perceived effort and frustration. This spike occurred despite some course runs that had similar mission performance even without any operator involvement. However, unpredictable scenarios without operator intervention did result in more error-related events. To assist in reducing operator load while maintaining the benefits of operator input, better communication between humans and robots is needed. This could include information exchange around what the robot team is finding difficult at present, and what their plans are to address it as a way to reduce operator intervention and cognitive load to resolve these issues~\cite{DEMIR2020102436,9563095,10.1145/3382507.3418871}. Alternatively, human operators could receive further practical training to reduce intervention frequency, building a greater sense of trust that the robot autonomy team will contribute to the goal in a beneficial way. Further improvements to human-robot team performance may also come from passive and/or continued exposure to the capability of the autonomous robot team, as well as further refinement of individual operator style to minimise their need to intervene, such as directing robots away from high risk areas well ahead of time. More operational improvements may also arise from new interfaces and systems being integrated into each mission run, given that human-robot collaboration research has shown the impact of natural language, shared cognition~\cite{DEMIR2020102436}, interaction dynamics~\cite{9563095}, emotion-led statements and affective expression to create effective human-robot teamwork~\cite{10.1145/3382507.3418871}. Therefore, the advantageous benefits of operator intervention might be better paired with new information exchange strategies that helps to reduce cognitive load, while at the same time, retaining the benefits of operator involvement. 

Operators reported that after viewing the fully autonomous runs, they increased their estimation of robot autonomy, including being less likely to intervene as often in future, showing greater trust in robot capabilities to complete the mission~\cite{9515476,8982042}. Transparency was reported to be an important factor in future human-robot team improvements, with requests that the robot team communicate its view of the world and its plans more effectively~\cite{8982042,9282970}. The operators had extensive experience with the system, mission objectives, and robotic hardware, but still reported that they were, to some extent, still learning and creating an accurate perception of the system's capability when viewing its use in different scenarios. This finding demonstrated that the limited communication of error states or robot intent may have resulted in operators intervening when they might not have needed to intervene to achieve the same level of performance. While autonomous behaviour may have improved team performance, this could have also contributed to a weakened state of situational awareness~\cite{doi:10.1177/1555343411409323}. 

\subsection{Strengths and Limitations} 
The experimental set has both strengths and limitations that should be noted, given the real-world nature of the deployment. Human operators involved in the missions had high levels of experience in the mission scenario, meaning that operators were familiar with the prospective bounds of the possible course layouts, and possible places where artifacts may have been hidden. Prior experience in a similar test site also meant that the operators could have changed their actions based on prior knowledge and experience to optimise their score. To conduct the mission runs, operators needed to be familiar with the system to operate it, and the difficulty level did not allow for less experienced staff to lead the missions. However, experienced operators provided a new insight into how human intervention can improve team-related outcomes, and interview data provided clearer insight into key variables relevant to expert operators. Experienced operators also provided insights about further features operators could use in the future. It also allowed operators to reflect on their performance in relation to previous missions, meaning that more skilled, nuanced and critical factors of human-robot team operations could be explored, especially compared to novice operators. Experienced operators were also required to run a more realistic scenario in the real world with physical robots, providing new challenges and scenarios not present in simulation. Furthermore, operators were blinded from knowing critical information about the experiment ahead of time, but it was not possible to completely ensure that no course information was known prior to the run. For instance, the course was required to be set up prior to the operator attending the demountable building. Operators were not permitted to discuss information between the team prior to experimental testing, and an external experimenter was present at all times to observe the changeover of operators, and to assist in quality control of the experiment. While some runs were inadvertently impacted by factors outside of the control of the experimenter and operators, these factors again represented more closely what can happen in real-life deployments, such as robots failing to start, or being separated from the operator station. While the real world nature of the experiment was more closely representative to a search and rescue mission compared to simulation experiments, the experiment required a standard set of artifacts and standardized course covering. While this reduces some of the realism, the design was intended to maintain similar tasks and principles that would otherwise be seen in a search and rescue mission with minimal-to-no additional robot help from safety observers and mission-run data collected as it was encountered on the field for each run. The design also only tested two human-machine team configurations, and other configurations and task-load allocation could be considered in future tests, such as mixed composition teams with multiple operators ~\cite{9474953} with a shared-pool of robots~\cite{doi:10.1177/1555343411409323}, or more closely testing different workload components related to the mission~\cite{8169072}. Furthermore, while the travelled distance did show fewer gains at the end of the mission time, operators did not appear to show task complacency or disengagement, representing a calibrated level of robot autonomy, task involvement, and course length to achieve the experimental outcome~\cite{doi:10.1080/00140139.2018.1441449}. 

\section{Conclusion}
A total of 16 real-world missions found that human-robot team operators do create notable advantages to search and rescue missions when paired with state-of-the-art robot autonomy,  compared to robot autonomy alone. Operators contribute to improved mission-based outcomes, help to overcome challenges that robots encounter during course runs that can impede progress, and help to recover robots faster out of scenarios that could lead to detrimental outcomes. For now, human-robot teams for search and rescue continue to be the recommended construct for covering the most ground, and helping to find the most items or victims in the shortest time possible. However, operators engaged in robot supervision are slower on average to review the information provided by the robots to determine a response, and so future consideration must be made to determine if the operators should either control fewer robots to help increase review and response time during the mission, or to forfeit their involvement in other human-robot leadership and control tasks to instead offer rapid response times to reach more global mission-related goals.  

\begin{acks}
  We would like to thank the technical team who assisted in monitoring robot safety during each course run, and the research students who assisted with data preparation. We would like to thank the CSIRO Data61 team who contributed to the development of the final system solution for the SubT Challenge from which this experiment was based on. We would like to thank the Collaborative Intelligence Future Science Platform (CINTEL FSP) for supporting this experiment. We would also like to acknowledge the support from other robot operators that assisted with this experiment: Mark Cox, Pavan Sikka, Md Komol, Tom Hines, Tirtha Bandyopadhyay, and Ben Tam. We would also like to acknowledge the help and support from Tom Hines for his assistance with the scoring server. 
\end{acks}

\bibliographystyle{ieeetr}
\bibliography{software}

\begin{thebibliography}{10}

\bibitem{bauer2008human}
A.~Bauer, D.~Wollherr, and M.~Buss, ``Human--robot collaboration: a survey,''
  {\em International Journal of Humanoid Robotics}, vol.~5, no.~01, pp.~47--66,
  2008.

\bibitem{doi:10.1080/00140139.2018.1441449}
J.~L. Wright, J.~Y.~C. Chen, and M.~J. Barnes, ``Human–automation interaction
  for multiple robot control: the effect of varying automation assistance and
  individual differences on operator performance,'' {\em Ergonomics}, vol.~61,
  no.~8, pp.~1033--1045, 2018.
\newblock PMID: 29451105.

\bibitem{9484733}
H.~Wu, A.~Ghadami, A.~E. Bayrak, J.~M. Smereka, and B.~I. Epureanu, ``Impact of
  heterogeneity and risk aversion on task allocation in multi-agent teams,''
  {\em IEEE Robotics and Automation Letters}, vol.~6, no.~4, pp.~7065--7072,
  2021.

\bibitem{Murphy2016}
R.~R. Murphy, S.~Tadokoro, and A.~Kleiner, {\em Disaster Robotics},
  pp.~1577--1604.
\newblock Cham: Springer International Publishing, 2016.

\bibitem{michaelis2020collaborative}
J.~E. Michaelis, A.~Siebert-Evenstone, D.~W. Shaffer, and B.~Mutlu,
  ``Collaborative or simply uncaged? understanding human-cobot interactions in
  automation,'' in {\em Proceedings of the 2020 CHI Conference on Human Factors
  in Computing Systems}, pp.~1--12, 2020.

\bibitem{DBLP:journals/corr/abs-2104-09053}
N.~Hudson, F.~Talbot, M.~Cox, J.~L. Williams, T.~Hines, A.~Pitt, B.~Wood,
  D.~Frousheger, K.~L. Surdo, T.~Molnar, R.~Steindl, M.~Wildie, I.~Sa,
  N.~Kottege, K.~Stepanas, E.~Hern{\'{a}}ndez, G.~Catt, W.~Docherty, B.~Tidd,
  B.~Tam, S.~Murrell, M.~Bessell, L.~Hanson, L.~Tychsen{-}Smith, H.~Suzuki,
  L.~Overs, F.~Kendoul, G.~Wagner, D.~Palmer, P.~Milani, M.~O'Brien, S.~Jiang,
  S.~Chen, and R.~C. Arkin, ``Heterogeneous ground and air platforms,
  homogeneous sensing: Team {CSIRO} {D}ata61's approach to the {DARPA}
  subterranean challenge,'' {\em Field Robotics}, vol.~2, 2022.

\bibitem{6697830}
J.~Y.~C. Chen and M.~J. Barnes, ``Human–agent teaming for multirobot control:
  A review of human factors issues,'' {\em IEEE Transactions on Human-Machine
  Systems}, vol.~44, no.~1, pp.~13--29, 2014.

\bibitem{doi:10.1177/1541931215591051}
C.~E. Bartlett and N.~J. Cooke, ``Human-robot teaming in urban search and
  rescue,'' {\em Proceedings of the Human Factors and Ergonomics Society Annual
  Meeting}, vol.~59, no.~1, pp.~250--254, 2015.

\bibitem{moonlight}
J.~L. Burke, R.~R. Murphy, M.~D. Coovert, and D.~L. Riddle, ``Moonlight in
  miami: Field study of human-robot interaction in the context of an urban
  search and rescue disaster response training exercise,'' {\em
  Human–Computer Interaction}, vol.~19, no.~1-2, pp.~85--116, 2004.

\bibitem{hambuchen2021review}
K.~Hambuchen, J.~Marquez, and T.~Fong, ``A review of nasa human-robot
  interaction in space,'' {\em Current Robotics Reports}, vol.~2, no.~3,
  pp.~265--272, 2021.

\bibitem{liu2013robotic}
Y.~Liu and G.~Nejat, ``Robotic urban search and rescue: A survey from the
  control perspective,'' {\em Journal of Intelligent \& Robotic Systems},
  vol.~72, no.~2, pp.~147--165, 2013.

\bibitem{vagia2016literature}
M.~Vagia, A.~A. Transeth, and S.~A. Fjerdingen, ``A literature review on the
  levels of automation during the years. what are the different taxonomies that
  have been proposed?,'' {\em Applied ergonomics}, vol.~53, pp.~190--202, 2016.

\bibitem{lewis2010choosing}
M.~Lewis, H.~Wang, S.~Y. Chien, P.~Velagapudi, P.~Scerri, and K.~Sycara,
  ``Choosing autonomy modes for multirobot search,'' {\em Human Factors},
  vol.~52, no.~2, pp.~225--233, 2010.

\bibitem{8169072}
B.~L. Hooey, D.~B. Kaber, J.~A. Adams, T.~W. Fong, and B.~F. Gore, ``The
  underpinnings of workload in unmanned vehicle systems,'' {\em IEEE
  Transactions on Human-Machine Systems}, vol.~48, no.~5, pp.~452--467, 2018.

\bibitem{4651073}
P.~Velagapudi, P.~Scerri, K.~Sycara, H.~Wang, M.~Lewis, and J.~Wang, ``Scaling
  effects in multi-robot control,'' in {\em 2008 IEEE/RSJ International
  Conference on Intelligent Robots and Systems}, pp.~2121--2126, 2008.

\bibitem{doi:10.1177/0018720817743223}
N.~J. McNeese, M.~Demir, N.~J. Cooke, and C.~Myers, ``Teaming with a synthetic
  teammate: Insights into human-autonomy teaming,'' {\em Human Factors},
  vol.~60, no.~2, pp.~262--273, 2018.
\newblock PMID: 29185818.

\bibitem{1307409}
J.~Scholtz, J.~Young, J.~Drury, and H.~Yanco, ``Evaluation of human-robot
  interaction awareness in search and rescue,'' in {\em IEEE International
  Conference on Robotics and Automation, 2004. Proceedings. ICRA '04. 2004},
  vol.~3, pp.~2327--2332 Vol.3, 2004.

\bibitem{6005237}
B.~Larochelle, G.-J.~M. Kruijff, N.~Smets, T.~Mioch, and P.~Groenewegen,
  ``Establishing human situation awareness using a multi-modal operator control
  unit in an urban search and rescue human-robot team,'' in {\em 2011 RO-MAN},
  pp.~229--234, 2011.

\bibitem{6926363}
T.~R. Colin, N.~J. Smets, T.~Mioch, and M.~A. Neerincx, ``Real time modeling of
  the cognitive load of an urban search and rescue robot operator,'' in {\em
  The 23rd IEEE International Symposium on Robot and Human Interactive
  Communication}, pp.~874--879, 2014.

\bibitem{doi:10.1177/1555343411409323}
M.~Lewis, H.~Wang, S.~Y. Chien, P.~Velagapudi, P.~Scerri, and K.~Sycara,
  ``Process and performance in human-robot teams,'' {\em Journal of Cognitive
  Engineering and Decision Making}, vol.~5, no.~2, pp.~186--208, 2011.

\bibitem{murphy2014disaster}
R.~R. Murphy, {\em Disaster robotics}.
\newblock MIT press, 2014.

\bibitem{8913876}
J.~Humann and K.~A. Pollard, ``Human factors in the scalability of multirobot
  operation: A review and simulation,'' in {\em 2019 IEEE International
  Conference on Systems, Man and Cybernetics (SMC)}, pp.~700--707, 2019.

\bibitem{doi:10.1177/154193120504900347}
R.~R. Murphy and J.~L. Burke, ``Up from the rubble: Lessons learned about hri
  from search and rescue,'' {\em Proceedings of the Human Factors and
  Ergonomics Society Annual Meeting}, vol.~49, no.~3, pp.~437--441, 2005.

\bibitem{yancodarpa2013}
H.~A. Yanco, A.~Norton, W.~Ober, D.~Shane, A.~Skinner, and J.~Vice, ``Analysis
  of human-robot interaction at the darpa robotics challenge trials,'' {\em
  Journal of Field Robotics}, vol.~32, no.~3, pp.~420--444, 2015.

\bibitem{10.1145/1349822.1349825}
J.~Wang and M.~Lewis, ``Assessing cooperation in human control of heterogeneous
  robots,'' in {\em Proceedings of the 3rd ACM/IEEE International Conference on
  Human Robot Interaction}, HRI '08, (New York, NY, USA), p.~9–16,
  Association for Computing Machinery, 2008.

\bibitem{6251723}
J.~Wang and M.~Lewis, ``Human control for cooperating robot teams,'' in {\em
  2007 2nd ACM/IEEE International Conference on Human-Robot Interaction (HRI)},
  pp.~9--16, 2007.

\bibitem{hong2019investigating}
A.~Hong, O.~Igharoro, Y.~Liu, F.~Niroui, G.~Nejat, and B.~Benhabib,
  ``Investigating human-robot teams for learning-based semi-autonomous control
  in urban search and rescue environments,'' {\em Journal of Intelligent \&
  Robotic Systems}, vol.~94, no.~3, pp.~669--686, 2019.

\bibitem{chen_multimodal_2022}
S.~Chen, M.~J. O'Brien, F.~Talbot, J.~Williams, B.~Tidd, A.~Pitt, and R.~Arkin,
  ``Multimodal user interface for multi-robot control in underground
  environments,'' in {\em {IEEE} {International} {Conference} on {Intelligent}
  {Robots} and {Systems}}, 2022.

\bibitem{Darparules}
D.~S. Challenge, ``Competition rules final event revision 1,'' 2021.

\bibitem{8404030}
P.~Damacharla, A.~Y. Javaid, J.~J. Gallimore, and V.~K. Devabhaktuni, ``Common
  metrics to benchmark human-machine teams (hmt): A review,'' {\em IEEE
  Access}, vol.~6, pp.~38637--38655, 2018.

\bibitem{elo2008qualitative}
S.~Elo and H.~Kyng{\"a}s, ``The qualitative content analysis process,'' {\em
  Journal of advanced nursing}, vol.~62, no.~1, pp.~107--115, 2008.

\bibitem{hsieh2005three}
H.-F. Hsieh and S.~E. Shannon, ``Three approaches to qualitative content
  analysis,'' {\em Qualitative health research}, vol.~15, no.~9,
  pp.~1277--1288, 2005.

\bibitem{vaismoradi2013content}
M.~Vaismoradi, H.~Turunen, and T.~Bondas, ``Content analysis and thematic
  analysis: Implications for conducting a qualitative descriptive study,'' {\em
  Nursing \& health sciences}, vol.~15, no.~3, pp.~398--405, 2013.

\bibitem{DEMIR2020102436}
M.~Demir, N.~J. McNeese, and N.~J. Cooke, ``Understanding human-robot teams in
  light of all-human teams: Aspects of team interaction and shared cognition,''
  {\em International Journal of Human-Computer Studies}, vol.~140, p.~102436,
  2020.

\bibitem{9563095}
M.~Demir, N.~J. McNeese, J.~C. Gorman, N.~J. Cooke, C.~W. Myers, and D.~A.
  Grimm, ``Exploration of teammate trust and interaction dynamics in
  human-autonomy teaming,'' {\em IEEE Transactions on Human-Machine Systems},
  vol.~51, no.~6, pp.~696--705, 2021.

\bibitem{10.1145/3382507.3418871}
S.~A. Akgun, M.~Ghafurian, M.~Crowley, and K.~Dautenhahn, ``Using emotions to
  complement multi-modal human-robot interaction in urban search and rescue
  scenarios,'' in {\em Proceedings of the 2020 International Conference on
  Multimodal Interaction}, ICMI '20, (New York, NY, USA), p.~575–584,
  Association for Computing Machinery, 2020.

\bibitem{9515476}
M.~Chiou, F.~McCabe, M.~Grigoriou, and R.~Stolkin, ``Trust, shared
  understanding and locus of control in mixed-initiative robotic systems,'' in
  {\em 2021 30th IEEE International Conference on Robot \& Human Interactive
  Communication (RO-MAN)}, pp.~684--691, 2021.

\bibitem{8982042}
A.~Bhaskara, M.~Skinner, and S.~Loft, ``Agent transparency: A review of current
  theory and evidence,'' {\em IEEE Transactions on Human-Machine Systems},
  vol.~50, no.~3, pp.~215--224, 2020.

\bibitem{9282970}
F.~Rajabiyazdi and G.~A. Jamieson, ``A review of transparency (seeing-into)
  models,'' in {\em 2020 IEEE International Conference on Systems, Man, and
  Cybernetics (SMC)}, pp.~302--308, 2020.

\bibitem{9474953}
N.~J. McNeese, B.~G. Schelble, L.~B. Canonico, and M.~Demir, ``Who/what is my
  teammate? team composition considerations in human–ai teaming,'' {\em IEEE
  Transactions on Human-Machine Systems}, vol.~51, no.~4, pp.~288--299, 2021.

\end{thebibliography}

  \begin{figure}
  \includegraphics[width=\linewidth]{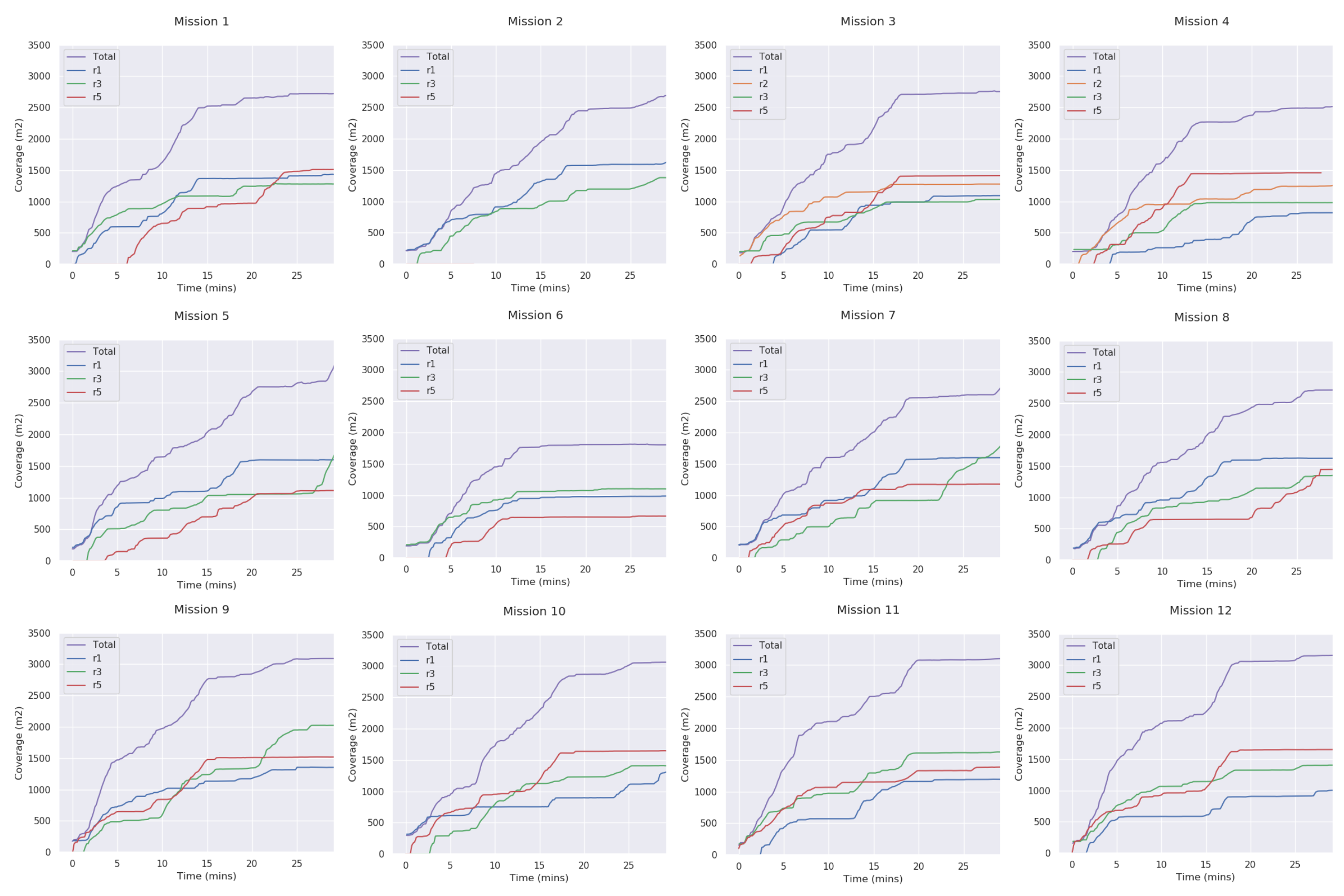}
  \caption{All Mission Courses - All Robots for Total Coverage in m$^{2}$ vs time. R1 is ATR 1, R2 is Spot 2, R3 is ATR 2, and R5 is Spot 1.}
  \label{fig:combinedcover}
  \end{figure}

  \begin{figure}
  \includegraphics[width=\linewidth]{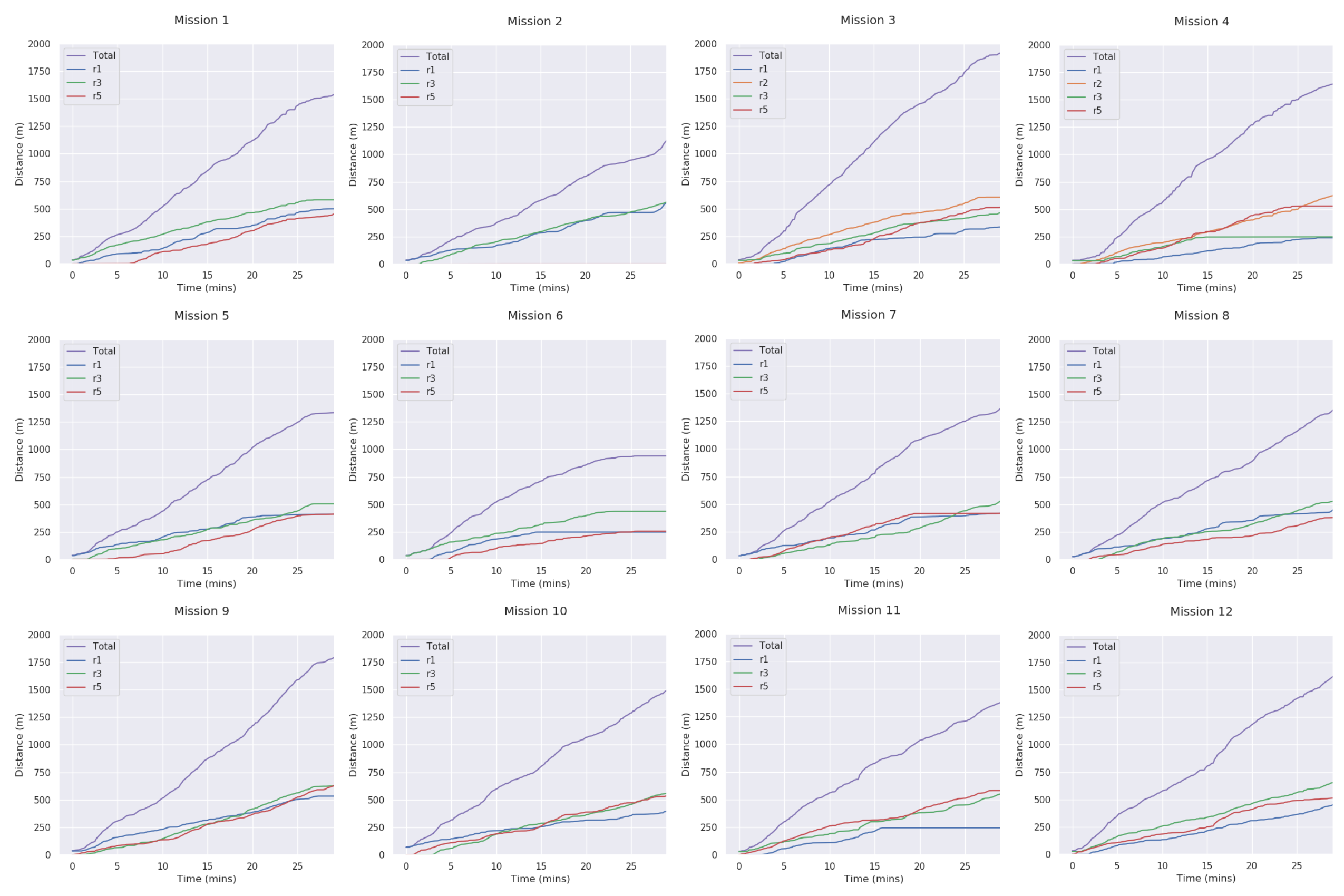}
    \caption{All Mission Courses - All Robots for Total Trajectory Distance in metres vs time. R1 is ATR 1, R2 is Spot 2, R3 is ATR 2, and R5 is Spot 1.}
  \label{fig:combinedtraj}
  \end{figure}

\end{document}